\documentclass[review,3p,times,10pt]{elsarticle} %
\usepackage{hyperref}
\usepackage{amsmath}
\usepackage{amsfonts}
\usepackage{graphicx}
\usepackage{algorithm}
\usepackage{algorithmicx}
\usepackage{algpseudocode}
\usepackage{booktabs}
\usepackage{array}
\usepackage{multirow}
\usepackage{xcolor}
\usepackage{colortbl}
\usepackage{titlesec}
\usepackage{epsfig}
\usepackage{epstopdf}
\usepackage{caption}
\usepackage{footmisc}
\usepackage{setspace}
\usepackage{txfonts}
\journal{Knowledge-Based SYSTEMS}
\begin{document}
\begin{frontmatter}
\title{High Performance Visual Tracking with Circular and Structural Operators}
\author[hitsz]{Peng Gao}
\author[hitsz]{Yipeng Ma}
\author[hitsz]{Ke Song}
\author[hitsz]{Chao Li}
\author[hitsz]{Fei Wang\corref{corresponding}}\ead{wangfeiz@hit.edu.cn}
\author[hitsz]{Liyi Xiao\corref{corresponding}}\ead{xiaoly@hit.edu.cn}
\author[hitsz]{Yan Zhang}
\address[hitsz]{Shenzhen Graduate School, Harbin Institute of Technology, China}
\cortext[corresponding]{Corresponding authors}
\begin{abstract}
Visual tracking algorithms based on structured output support vector machine (SOSVM) have demonstrated excellent performance. However, sampling methods and optimization strategies of SOSVM undesirably increase the computational overloads, which hinder real-time application of these algorithms. Moreover, due to the lack of high-dimensional features and dense training samples, SOSVM-based algorithms are unstable to deal with various challenging scenarios, such as occlusions and scale variations. Recently, visual tracking algorithms based on discriminative correlation filters (DCF), especially the combination of DCF and features from deep convolutional neural networks (CNN), have been successfully applied to visual tracking, and attains surprisingly good performance on recent benchmarks. The success is mainly attributed to two aspects: the circular correlation properties of DCF and the powerful representation capabilities of CNN features. Nevertheless, compared with SOSVM, DCF-based algorithms are restricted to simple ridge regression which has a weaker discriminative ability. In this paper, a novel circular and structural operator tracker (CSOT) is proposed for high performance visual tracking, it not only possesses the powerful discriminative capability of SOSVM but also efficiently inherits the superior computational efficiency of DCF. Based on the proposed circular and structural operators, a set of primal confidence score maps can be obtained by circular correlating feature maps with their corresponding structural correlation filters. Furthermore, an implicit interpolation is applied to convert the multi-resolution feature maps to the continuous domain and make all primal confidence score maps have the same spatial resolution. Then, we exploit an efficient ensemble post-processor based on relative entropy, which can coalesce primal confidence score maps and create an optimal confidence score map for more accurate localization. The target is localized on the peak of the optimal confidence score map. Besides, we introduce a collaborative optimization strategy to update circular and structural operators by iteratively training structural correlation filters, which significantly reduces computational complexity and improves robustness. Experimental results demonstrate that our approach achieves state-of-the-art performance in mean AUC scores of 71.5\% and 69.4\% on the OTB2013 and OTB2015 benchmarks respectively, and obtains a third-best expected average overlap (EAO) score of 29.8\% on the VOT2017 benchmark.
\end{abstract}
\end{frontmatter}
%
%\begin{spacing}{2.0}
%%
%\linenumbers
%%
\section{Introduction} \label{sec:intro}

\indent Visual tracking is one of the fundamental research problems in the field of computer vision, it has a variety of applications such as video surveillance, assistant driving systems, service robots and human-computer interaction. Given the initial position and size of an arbitrary target, a high performance tracker should discriminate the target from the background and locate the target in subsequent frames~\cite{winsty2015}. Despite significant progress have been made in the past decade, there are still numerous unsolved visual tracking challenges due to various negative factors, such as motion blur, illumination variations, deformations and occlusions.

\indent In general, visual tracking algorithms can be categorized as discriminative or generative approaches. Generative tracking algorithms~\cite{survey2014,asms,hnt,dat} establish a reference appearance model to characterize the target in current frame, and then an image patch most similar to the reference model is searched as the target in the next frame. The searching process can be guided by a probabilistic motion model. However, the background information contained in the initial target bounding box is also modeled, which weakens the similarity measurement and thus limits the performance of generative tracking algorithms. In contrast, discriminative tracking algorithms have received unprecedented research interest in the last decade. Most discriminative algorithms follow the tracking-by-detection paradigm~\cite{tld,struck,stc,kcf,ca,ma2018visual}, which treats the tracking task as a detection problem. They employ a classifier or a regressor to process both target and background representations, and produce an optimal decision boundary that can efficiently discriminate the target from the background.

\indent Support vector machine (SVM) is a popular representative and discriminative appearance classifier~\cite{svt,meem,scf}. Existing SVM-based trackers mostly have two modules: a sampler and a learner. The sampler generates a set of positive and negative training samples, and the learner updates the classifier based on training samples. SOSVM-based trackers outperform other SVM-based trackers considerably~\cite{otb2013,struck}. However, using dense training samples and high dimensional features increases computational complexity, therefore limiting the performance of these SOSVM-based trackers, and the optimization and detection process may further degrade the computational speed~\cite{struck,dlssvm}.

\indent Recently, discriminative correlation filters (DCF) based trackers~\cite{ecs,cnt,muster,mcft,mfcmt} have achieved excellent results in terms of accuracy, robustness and speed~\cite{otb2013,otb2015,vot2017}, because they treat tracking tasks as similarity learning problems. Since DCF exploits all circular shifts of training samples to solve a ridge regression in the Fourier frequency domain~\cite{kcf}, it avoids time-consuming correlation operations. Recent advancements in DCF-based tracking performance are driven by multi-dimensional features~\cite{kcf,cnt}, adaptive scale estimation~\cite{samf,dsst}, robust long-term memory components~\cite{ltc,ltst}, reducing boundary effects~\cite{bacf,srdcf} and continuous-domain features~\cite{ccot,eco}. Most DCF-based trackers employ conventional handcrafted appearance features, such as raw pixels~\cite{mosse}, Histograms of Oriented Gradients~(HOG)~\cite{hog,kcf}, ColorNames~(CN)~\cite{cn,cnt} or their combinations~\cite{staple,ieice}. Nevertheless, compared with SOSVM-based trackers, DCF-based trackers are restricted to simple ridge regression which has a weaker discriminative ability.

\indent Due to their powerful representation capabilities, deep CNN features have been successfully applied to many computer vision tasks, including image classification~\cite{alexnet,vgg} and object detection~\cite{Girshick2013Rich,Liu2015SSD}. A CNN usually composes of several layers of convolution, local normalization and polling operations. Different layers capture different levels of features. Some recent works combine DCF with deep features have achieved state-of-the-art performance on many tracking benchmarks~\cite{vot2016,vot2017}. Two most successful deep CNN features extracted from pre-trained networks are deep appearance features~\cite{deepsrdcf} and deep motion features~\cite{deepmotion}. Despite deep CNN features have strong capacities to represent targets, extracting them requires intensive computations.

\indent To deal with challenging scenarios such as occlusions and deformations, several recent approaches~\cite{imt,mtmvt,branchout,Li2016Convolutional} exploit ensemble methods to combine multiple different types of features, models or trackers. Many algorithms take such ensemble strategies, including interactive Markov Chain Monte Carlo~\cite{Kwon}, factorial hidden Markov model~\cite{ebt}, entropy minimization~\cite{meem}, random ferns~\cite{Rao2012Online}, sharing convolutional layers~\cite{Nam2016Modeling} and linear interpolation~\cite{staple}. Although the ensemble method is effective and robust for visual tracking, it still suffers from high computational cost and thus has not been applied widely.
\begin{figure}[t]
\captionsetup{belowskip=-0em}
\centering
\begin{minipage}[b]{\linewidth}\centerline{\includegraphics[width=\textwidth]{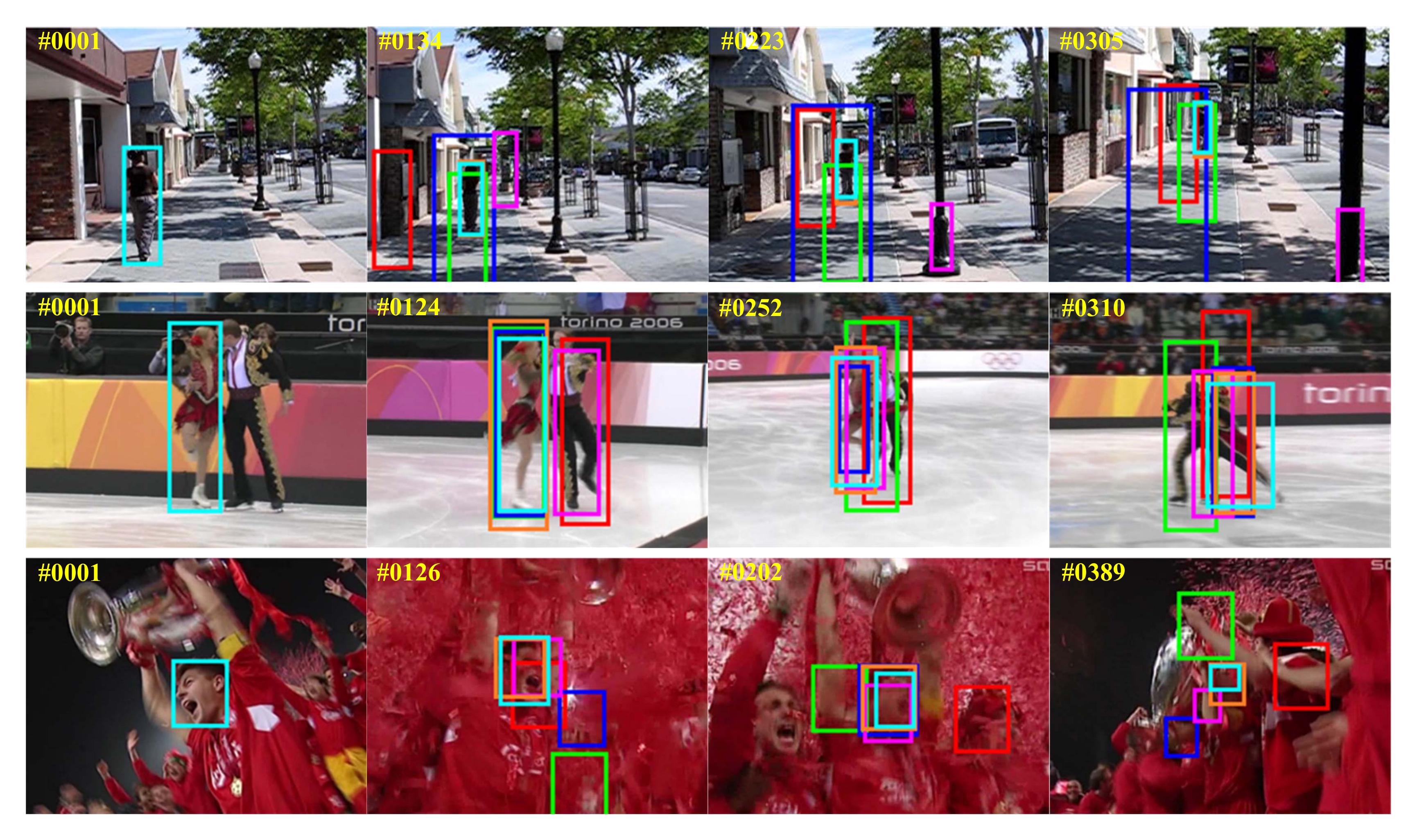}}\end{minipage}
\begin{minipage}[b]{0.7\linewidth}\centerline{\includegraphics[width=\textwidth]{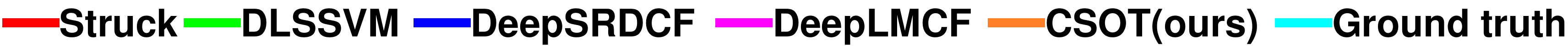}}\medskip\end{minipage}
\vspace{-1em}
\caption{Comparison of our method CSOT~(in~\textcolor[rgb]{1.00,0.50,0.15}{orange})~with four state-of-the-art trackers Struck~\cite{struck}~(in~\textcolor[rgb]{1.00,0.00,0.00}{red}), DLSSVM~\cite{dlssvm}~(in~\textcolor[rgb]{0.00,1.00,0.00}{green}), DeepSRDCF~\cite{deepsrdcf}~(in~\textcolor[rgb]{0.00,0.00,1.00}{blue})~and DeepLMCF~\cite{lmcf}~(in~\textcolor[rgb]{1.00,0.00,1.00}{pink})~on three example sequences~\emph{Human9}~(top row),~\emph{Skating}~(middle row) and~\emph{Soccer}~(bottom row) from OTB2015 benchmark~\cite{otb2015}, respectively. The ground-truth bounding boxes of these sequences are also given~(in~\textcolor[rgb]{0.00,1.00,1.00}{cyan}). Best viewed in color.}%
\label{fig:1}
\end{figure}

\indent In this paper, a novel circular and structural operator based tracker (CSOT) is proposed by incorporating DCF and SOSVM, which not only possesses the powerful discriminative capability of SOSVM~\cite{struck} but also efficiently inherits the superior computational efficiency of DCF~\cite{kcf}. So that it can employ higher-dimensional deep features and denser circular shifted samples. We employ deep complementary features including both deep appearance features and deep motion features to enhance the diversity of training samples. Similar to \cite{deepmotion}, the shallow (\emph{Conv}-1) and deep (\emph{Conv}-5) convolutional layers of the \emph{VGG-M}~\cite{vgg} network are exploited to extract deep appearance feature maps, and they have 128 and 512 feature channels respectively. Deep motion feature maps are extracted from the \emph{Conv}-5 layer of the pre-trained optical flow network as described in \emph{ActionCubes}~\cite{actiontube}~\footnote{\url{https://gkioxari.github.io/ActionTubes}}, which has 384 feature channels. In CSOT, a set of primal confidence score maps can be obtained by circular correlating deep complementary feature maps with their corresponding structural correlation filters using circular and structural operators. Moreover, we have observed the fact that some approaches merely sum up all primal confidence score maps to estimate target positions. Unfortunately, they are not reliable for visual tracking. To address this issue, we introduce an ensemble post-processor based on relative entropy, which can coalesce primal confidence score maps and create an optimal confidence score map for more accurate localization. The target can be localized by searching the peak value on the optimal confidence score map. However, the ensemble post-processor requires all primal confidence score maps have the same spatial resolution, explicitly resampling all primal confidence score maps to the same resolution will introduce distortion and even decrease the tracking performance. Therefore, we exploit an implicit interpolation to extract continuous-domain deep feature maps and make all primal confidence score maps have the same spatial resolution~\cite{ccot,eco}. Besides, we introduce a collaborative optimization strategy to update circular and structural operators by iterative training structural correlation filters. To boost the tracking accuracy, we adopt a robust scale estimation for our CSOT. The main contributions of this paper can be summarized as follows:
\begin{itemize}
    \item A novel and efficient CSOT is proposed, which possesses the superior computational efficiency of DCF and the powerful discriminative capability of SOSVM. By employing circular and structural operators, continuous-domain deep feature maps and structural correlation filters are gathered to calculate primal confidence score maps with the same spatial resolution.
    \item An online collaborative optimization strategy is suggested to train structural correlation filters, which decomposes the training problem into two independent sub-problems and can be iteratively solved online.
    \item An ensemble post-processor is introduced, which combines primal confidence score maps to create an optimal confidence score map. Furthermore, a robust scale estimation scheme is adopted, which enables our tracker to adapt to target scale variations efficiently.
\end{itemize}

\indent Finally, experimental evaluations are conducted on several recent visual tracking benchmarks~\cite{otb2013,otb2015,vot2017}. As shown in Fig.~\ref{fig:1}, the results demonstrate that our tracker outperforms most state-of-the-art trackers both in terms of accuracy and robustness. To the best of our knowledge, we are the first to propose circular and structural operators for visual tracking with multiple deep complementary features in the continuous spatial domain.

\indent The rest of the paper is organized as follows. Section~\ref{sec:relat} briefly reviews related works. Section~\ref{sec:appro} details the proposed approach. Section~\ref{sec:exp} demonstrates the experimental results. Finally, Section~\ref{sec:con} concludes the paper.
\section{Related works} \label{sec:relat}
In this section, we give a brief review of related works in four categories: SVM-based trackers, DCF-based trackers, CNN-based trackers and ensemble-based trackers.

\subsection{SOSVM-based trackers}
SOSVM is a popular and successful backbone for visual tracking in the past decade~\cite{otb2013}. It treats visual tracking as a structured estimation that admits a consistent target representation for both optimization and detection. Struck~\cite{struck} is the first tracker employs SOSVM and demonstrates superior performance on the original OTB benchmark~(OTB2013)~\cite{otb2013}. But it is time-consuming and difficult to be extended to higher-dimensional features. DLSSVM~\cite{dlssvm} presents a dual linear SOSVM framework that approximates intersection kernels for feature representations based on explicit feature maps to improve tracking performance. Nonetheless, DLSSVM still suffers from high computational complexity, and it is hard to be used for real-time tracking. SCF~\cite{scf} derives an equivalent formulation of a SOSVM model with circulant samples to learn support correlation filters. LMCF~\cite{lmcf} exploits DCF to speed up SOSVM-based tracking models. Despite of their outstanding performances, employing high-dimensional features and dense training samples are computationally expensive, which make them difficult to be applied for real-time tracking.

\subsection{DCF-based trackers}
In recent years, DCF-based approaches have been successfully applied to visual object tracking. MOSSE~\cite{mosse} is the first tracker which exploits Fast Fourier Transform (FFT) to train single-channel correlation filters based on raw pixel samples of both the target and the background. Since MOSSE uses linear classifier and grayscale features, it has limited capability to track targets. To address this issue, Henriques \emph{et al}.~\cite{kcf} introduce a kernelized correlation filter-based approach~(KCF), which exploits kernel tricks and multi-dimensional HOG features for visual tracking. Subsequently, several DCF-based trackers are proposed to address the inherent limitations of DCF. Both SAMF~\cite{samf} and DSST~\cite{dsst} use scale pyramid representations to adaptively estimate scale variations of targets. Another deficiency of the DCF-based trackers is that since dense sampling strategies employ periodical training samples, thereby incurring unwanted boundary effects. It severely compromises the robustness of DCF-based tracking approaches, especially when targets are in the challenging scenario of out-of-view. To solve this problem, SRDCF~\cite{srdcf} adds a spatially regularized component to penalize DCF close to searching boundaries, while BACF~\cite{bacf} adopts a zero-padding scheme to ensure that correlation filters have the same size as sample patches. Some recent works~\cite{ccot,eco,cfwcr,lmsco} integrates multi-resolution features into formulations and learn a set of convolution filters to generate continuous-domain confidence score maps of the target. These trackers have shown excellent performance on recent visual tracking benchmarks~\cite{otb2015,vot2016,vot2017}.

\subsection{CNN-based trackers}
Because of their impressive representation power of deep features, some recent works combine them with DCF, and show state-of-the-art performance both in terms of accuracy and robustness. MDNet~\cite{mdnet} follows a paradigm of offline training and online fine-tuning. It has two purposes: (a) to learn domain-independent representations from pre-trained networks, and (b) to capture domain-specific information through online learning. Inspired by the Siamese network which has five convolutional layers, SiamFC~\cite{siamfc} introduces two identical branches, each branch has two conv5 layers. It then employs a new cross-correlation layer to connect the two conv5 layers together. Due to its high tracking speed, SiamFC is claimed to be the best real-time tracker in recent VOT challenge~\cite{vot2017}. However, most CNN-based trackers solely employ deep appearance features, but lack of high-level motion cues. Based on SRDCF method~\cite{srdcf}, DMSRDCF~\cite{deepmotion} investigates the fusion of conventional handcrafted features, deep appearance features and deep motion features, and it outperforms SRDCF significantly. In general, due to their advantages both in robustness and computational efficiency, CNN-based trackers have shown that deep features are more suitable for visual tracking.

\subsection{Ensemble-based trackers}
Recent years have witnessed significant advances of ensemble-based methods on visual tracking. Kwon and Lee~\cite{Kwon} use principal component analysis to construct several basic models and exploit an interactive Markov Chain Monte Carlo method, which combines those basic models together. EBT~\cite{ebt} formulates an ensemble tracking framework based on a factorial hidden Markov model, which leverages structured crowdsourced time series data of five independent trackers. Hong \emph{et al}.~\cite{mtmvt} present an efficient tracking approach based on a multi-task and multi-view joint sparse method. They employ a particle filter framework to combine each feature view linearly. MEEM~\cite{meem} proposes an entropy-regularized restoration scheme to restore the historical trackers, and then selects estimations of these trackers based on the minimum entropy criterion to address undesirable model updates. Staple~\cite{staple} makes use of two complementary tracking models in a ridge regression framework and linearly fuses two different results. MFCMT~\cite{mfcmt} combines the response maps of DSST and DAT based on a simple relative entropy criterion. All ensemble-based trackers have shown superior performance compared with their basic models.

\indent C-COT~\cite{ccot} and SCF~\cite{scf} are two works most closely related to ours. They focus on learning continuous convolution filters and support correlation filters, respectively. Different from these works, our work mainly aims to bridge the gap between DCF and SOSVM by using a novel and efficient circular and structural operator for visual tracking. Both the two works employ deep appearance features and conventional handcrafted features. In contrast, our approach exploits both deep appearance features and deep motion features extracted from a pre-trained VGG network~\cite{vgg} and a pre-trained optical flow network~\cite{actiontube}, respectively. Furthermore, our approach exploits interpolation to build continuous-domain feature maps, while SCF uses merely the explicitly resampling strategy to make feature maps have the same resolution.
\section{The proposed approach} \label{sec:appro}
\begin{figure}[t]
\captionsetup{belowskip=-0em}
\begin{minipage}[b]{\linewidth}\centerline{\includegraphics[width=\textwidth]{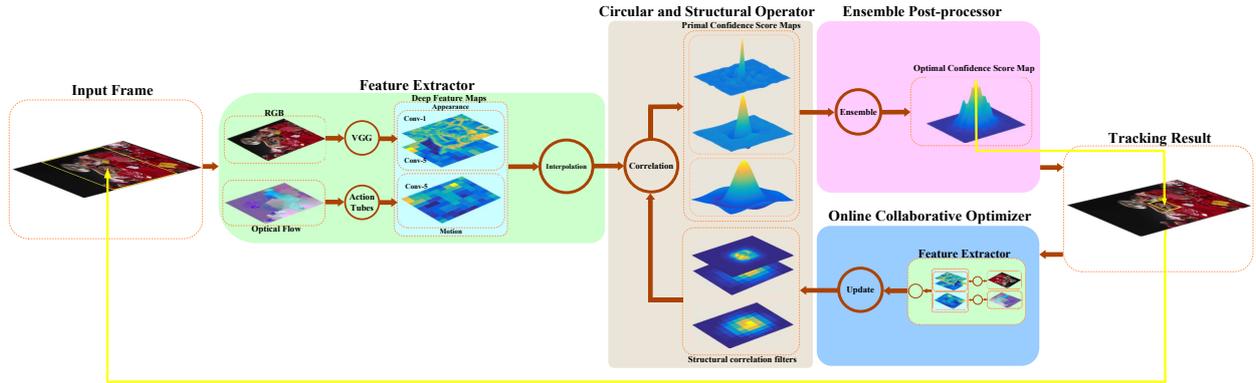}}\medskip\end{minipage}
\vspace{-2em}
\caption{The overall pipeline of our proposed approach.}
\label{fig:2}
\end{figure}
In this section, we describe our proposed CSOT in detail. The overall pipeline of CSOT is shown in Fig.~\ref{fig:2}, which is composed of four portions, i.e., the feature extractor, circular and structural operator, ensemble post-processor and online collaborative optimizer. In the pipeline, we firstly obtain searching samples and optical flows of the input image. Then in the feature extractor, the interpolated continuous deep complementary features of the target can be extracted from two independent deep networks. All these deep features are fed into the circular and structural operator to circular correlate with their corresponding structural correlation filters, primal confidence score maps with the same spatial resolution can be created. In the ensemble post-processor, all these primal confidence maps are combined to form a more discriminative optimal confidence score map. The estimated target size and position can be determined by searching the peak value on the optimal confidence score maps. Finally, we utilize the tracking results to update circular and structural operators by iterative training of structural correlation filters using the online collaborative optimizer.
\subsection{Circular and Structural Operators}\label{sec:appro1}
In our proposed approach, the aim is to learn the circular and structural operator $S_{\mathbf{w}}$ parameterized by structural correlation filters $\mathbf{w}$ from a set of circular input-output pairs $(\mathcal{X}_t,\mathcal{Y}_t)$, where $\mathcal{X}_t$ denotes an input space and $\mathcal{Y}_t$ indicates an arbitrary circular output space. For the sake of clarity, we formulate our approach for data defined in the one-dimensional domain in this section, and it can be generalized to higher dimensions in the same way~\cite{lmcf,ccot}. All circular shifted versions of the target image patch $\mathbf{x}_i$ of size $M_i\times N_i$ centered around the estimated target position $\mathbf{y}_{i}$ are considered as training samples~\cite{kcf} at frame $i$. Hence, the input space $\mathcal{X}_t$ contains the estimated target image patches, and the output space $\mathcal{Y}_t$ is now representing the estimated target positions on the circular shifted image patches,i.e. $\mathcal{Y}_t=\Big\{\mathbf{y}_{i(p)}\,|\,p\in\{0,\ldots,\Lambda-1\}\Big\}$, where $\Lambda=M_i\times N_i$ denotes the number of circular shifted target positions and $p$ indicates the relative pixel number of circular shift according to $\mathbf{y}_{i}$, e.g., $\mathbf{y}_{i(0)}$ denotes the estimated target position with no circular shift and $\mathbf{y}_{i(1)}$ is obtained by circular shifting $\mathbf{y}_{i}$ by 1 pixel. We use $(\mathcal{X}_t,\mathcal{Y}_t)$ to represent all training samples and $(\mathbf{x}_i,\mathbf{y}_{i(p)})$ denotes one training sample at frame $i$. The feature maps of these samples are denoted as $\varphi(\mathbf{x}_i,\mathbf{y}_{i(0)})\in\mathbb{R}^{\Lambda\times D}$, e.g., the deep feature map extracted from \emph{Conv-1} of the pre-trained \emph{vgg-m} network, $D$ indicates the number of feature map channels.

\indent Since $(\mathcal{X}_t,\mathcal{Y}_t)_i=\big[(\mathbf{x}_i,\mathbf{y}_{i(0)}),(\mathbf{x}_i,\mathbf{y}_{i(1)}),\ldots,(\mathbf{x}_i,\mathbf{y}_{i(\Lambda-1)})\big]$ is a circular matrix formed by enumerating all shifted training samples, we can calculate the primal confidence score map $S_i\in\mathbb{R}^{\Lambda}$ by using the circular and structural operator $S_{\mathbf{w}}$ to circular correlate structural correlation filters $\mathbf{w}$ with the feature map $\varphi(\mathbf{x}_i,\mathbf{y}_{i(0)})$,
\begin{equation} \label{eq:30}
\begin{aligned}
S_{i(\mathcal{Y}_t)}&=S_{\mathbf{w}}(\mathcal{X}_t,\mathcal{Y}_t)_i\\
&=\sum_{d=1}^D \Big(\varphi^d(\mathcal{X}_t,\mathcal{Y}_t)_i\Big)^T\mathbf{w}^d\\
&=\sum_{d=1}^D \mathbf{w}^d\scriptstyle\bigotimes\textstyle\varphi^d(\mathbf{x}_i,\mathbf{y}_{i(0)})
\end{aligned}
\end{equation}
where the symbol $\scriptstyle\bigotimes$ denotes the circular correlation, and $\mathbf{w}^d\in \mathbb{R}^{\Lambda}$ indicates the corresponding structural correlation filter of the $d$-th channel feature map $\varphi^d(\mathbf{x}_i,\mathbf{y}_{i(0)})$. The circular and structural operator can be learned by training structural correlation filters $\mathbf{w}$ following the optimization problem,
\begin{equation} \label{eq:2}
\begin{aligned}
&\min_{\mathbf{w}}{\sum_{d=1}^D\|\mathbf{w}^d\|^2+C\sum_{\mathbf{y}_{i(p)}\in\mathcal{Y}_t}{\xi^2_{i(\mathbf{y}_{i(p)})}}}\\
&\mathrm{s.t.}\;\forall p,\;\forall \mathbf{y}_{i(p)}\in\mathcal{Y}_t:\,\xi_{i(\mathbf{y}_{i(p)})}\geq J_{i(\mathbf{y}_{i(p)})}-
(S_{i(\mathbf{y}_{i(0)})}-S_{i(\mathbf{y}_{i(p)})})
\end{aligned}
\end{equation}
Here, $S_{i(\mathbf{y}_{i(p)})}=\sum_{d=1}^D \Big(\varphi^d(\mathbf{x}_i,\mathbf{y}_{i(p)}\Big)^T\mathbf{w}^d\in \mathbb{R}$ denotes the primal confidence score of each cyclic shifted training sample $(\mathbf{x}_i,\mathbf{y}_{i(p)})$, the slack variables $\xi_{i(p)}$ represents the penalty assigned to each sample for margin violations. For the desired training samples, $\xi_{i(p)}$ will be 0. The regularization parameter $C$ penalizes the complicated functions which are prone to over-fitting, and biases Eq.~\ref{eq:2} towards training error minimization and margin maximization. $J_{i(\mathbf{y}_{i(p)})}\in \mathbb{R}$ is the cost function which denotes the training error associated with $\mathbf{y}_{i(p)}$ and $\mathbf{y}_{i(0)}$. The cost function can be defined as,
\begin{equation} \label{eq:3}
J_{i(\mathbf{y}_{i(p)})}=1-m_{\mathbf{x}}(\mathbf{y}_{i(p)}),\;\; p=0,1,\ldots,\Lambda-1
\end{equation}
The desired confidence score output $m_{\mathbf{x}}(\mathbf{y}_{i(p)})$ is supposed to have a Gaussian form, i.e., $m_{\mathbf{x}}(t)=\frac{1}{\sqrt{2\pi\sigma^2}}\exp\Big(-\frac{(t-\mathbf{y}_{i(0)})^2}{2\sigma^2}\Big)$, which takes a peak value of 1 located at the estimated target position $\mathbf{y}_{i(0)}$ and smoothly reduces to 0 for larger circular shifts~\cite{kcf}.

\indent However, feature maps extracted from each convolutional layer of a deep network frequently have different resolution. In order to combine the primal confidence score for target detection in Section~\ref{sec:appro3}, all primal confidence score maps must have the same spatial resolution. To achieve this, we take advantage of an implicit interpolation to convert those feature maps to the continuous spatial domain. Suppose feature map $\varphi(\mathbf{x}_i,\mathbf{y}_{i(p)})$ extracted from a specific convolutional layer has an independent resolution $N$. Then, the feature map can be rewritten as a function $\varphi(\mathbf{x}_i,\mathbf{y}_{i(p)})[n]$ indexed by the discrete spatial variable $n\in\{0,1,\ldots,N-1\}$. Consequently, for each feature channel $d$, we convert the feature map from discrete spatial domain to the continuous spatial domain $t\in[0,\,T)$ by employing an interpolation operator $\Phi$,
\begin{equation} \label{eq:4}
\Phi^d\Big\{\varphi(\mathbf{x}_i,\mathbf{y}_{i(p)})\Big\}(t)=\sum_{n=0}^{N-1}\varphi^d(\mathbf{x}_i,\mathbf{y}_{i(p)})[n]b(t-\frac{T}{N}n)
\end{equation}
Here, $b$ represents the interpolation function with the period of $T>0$, and $\Phi\Big\{\varphi(\mathbf{x}_i,\mathbf{y}_{i(p)})\Big\}$ is the interpolated feature map, which viewed as a continuous $T$-periodic function. In our framework, the cubic spline kernel is applied to construct the interpolation function as described in C-COT~\cite{ccot}.

\indent Therefore, we can obtain primal confidence score maps with the same resolution by exploiting interpolated feature maps $\Phi\Big\{\varphi(\mathbf{x}_i,\mathbf{y}_{i(0)})\Big\}$ and structural correlation filters $\mathbf{w}$. Since the circular correlation is equivalent to an element-wise multiplication in the Fourier frequency domain, we can fast compute Eq.~\ref{eq:30} with the circulant property of circular matrix~\cite{Davis1979Circulant,Gray2005Toeplitz} as,
\begin{equation} \label{eq:5}
\begin{aligned}
S_{i(\mathcal{Y}_t)}=\mathcal{F}^{-1}\bigg(\sum_{d=1}^D\hat{\mathbf{w}}^d\scriptstyle\bigodot\textstyle
\hat{\Phi}^d\Big\{\varphi(\mathbf{x}_{i+1},\mathbf{y}_{i(0)})\Big\}\bigg)
\end{aligned}
\end{equation}
where the hat symbol $ \wedge $ of a $T$-periodic function represents the discrete Fourier transform (DFT), i.e., $\hat{\textbf{w}}^d[k]=\frac{1}{T}\int_0^T\textbf{w}^d(t)e^{-i\frac{2\pi k}{T}t}$, $\hat{\Phi}^d\Big\{\varphi(\mathcal{X}_t,\mathcal{Y}_t)_i\Big\}[k]=\sum_{n=0}^{N-1}\varphi^d(\mathcal{X}_t,\mathcal{Y}_t)_i[n]e^{-i\frac{2\pi k}{N}}\hat{b}[k]$ and $\hat{b}[k]=\frac{1}{N}\exp(-i\frac{\pi}{N})\hat{b}(\frac{k}{N})$. $\mathcal{F}^{-1}$ stands for the inverse discrete Fourier transform (IDFT), i.e., $S_{i(\mathcal{Y}_t)}(t)=\mathcal{F}^{-1}\Big(\hat{S}_{i(\mathcal{Y}_t)}[k]\Big)=\sum_{-\infty}^{+\infty}\hat{S}_{i(\mathcal{Y}_t)}[k]e^{i\frac{2\pi k}{T}t}$, $\hat{S}_{i(\mathcal{Y}_t)}[k]$ are Fourier coefficients which can be obtained as $\hat{S}_{i(\mathcal{Y}_t)}[k]=\sum_{d=1}^D\hat{\textbf{w}}^d[k]\hat{\Phi}^d\Big\{\varphi(\mathcal{X}_t,\mathcal{Y}_t)_i\Big\}[k]$. The symbol $\scriptstyle\bigodot$ denotes the element-wise multiplication.

\indent Similar to SRDCF~\cite{srdcf}, we add a penalty function $\boldsymbol{\gamma}$ in Eq.~\ref{eq:2} to mitigate drawbacks of the periodic assumption and control the spatial extent of structural correlation filters $\mathbf{w}$. The learning problem can be equivalently formulated as,
\begin{equation} \label{eq:6}
\begin{aligned}
&\min_{\mathbf{w}}{\sum_{d=1}^D\|\boldsymbol{\gamma}\scriptstyle\bigodot\textstyle\mathbf{w}^d\|^2+C\sum_{\mathbf{y}_{i(p)}\in\mathcal{Y}_t}{\xi^2_{i(\mathbf{y}_{i(p)})}}}\\
&\mathrm{s.t.}\;\forall p,\;\forall \mathbf{y}_{i(p)}\in\mathcal{Y}_t:\,\xi_{i(\mathbf{y}_{i(p)})}\geq J_{i(\mathbf{y}_{i(p)})}-
(S_{i(\mathbf{y}_{i(0)})}-S_{i(\mathbf{y}_{i(p)})})
\end{aligned}
\end{equation}
This penalty function $\boldsymbol{\gamma}$ not only makes structural correlation filters $\mathbf{w}$ reduce smoothly from the center of the sample to boundaries but also suppress $\mathbf{w}$ which resides close to boundaries.
\subsection{Online collaborative optimization}\label{sec:appro2}
In order to optimize Eq.~\ref{eq:6} efficiently, a new parameter $\boldsymbol{\epsilon}_{i(\mathcal{Y}_t)}\in \mathbb{R}^{\Lambda}$ is defined, where $\boldsymbol{\xi}_{i(\mathcal{Y}_t)}=\boldsymbol{\epsilon}_{i(\mathcal{Y}_t)}+J_{i(\mathcal{Y}_t)}-
(S_{i(\mathbf{y}_{i(0)})}-S_{i(\mathcal{Y}_t)}),\; \boldsymbol{\epsilon}_{i(\mathcal{Y}_t)}\geq0$. Therefore, the minimization of Eq.~\ref{eq:6} is equivalent to the following problem,
\begin{equation} \label{eq:7}
\begin{aligned}
&\min_{\mathbf{w}}\sum_{d=1}^D\|\boldsymbol{\gamma}\scriptstyle\bigodot\textstyle\mathbf{w}^d\|^2+C\|\boldsymbol{\epsilon}_{i(\mathcal{Y}_t)}+J_{i(\mathcal{Y}_t)}-
(S_{i(\mathbf{y}_{i(0)})}-S_{i(\mathcal{Y}_t)})\|_2^2\\
&\mathrm{s.t.}\;\boldsymbol{\epsilon}_{i(\mathcal{Y}_t)}\geq0
\end{aligned}
\end{equation}

\indent There are two variables $\boldsymbol{\epsilon}_{i(\mathcal{Y}_t)}$ and $\mathbf{w}$ in this formulation have to be solved. When $\mathbf{w}$ is known, the subproblem on $\boldsymbol{\epsilon}_{i(\mathcal{Y}_t)}$ has a closed-form solution. However, when $\boldsymbol{\epsilon}_{i(\mathcal{Y}_t)}$ is known, the subproblem on $\mathbf{w}$ does not have the closed-form solution, since we exploit the spatial regularization and implicit interpolation. We employ the Conjugate Gradient~(CG) to iteratively solve the subproblem on $\mathbf{w}$. Motivated by SCF~\cite{scf}, we propose an online collaborative optimization that solves these problems efficiently by iterating between the following two steps.

\indent \textbf{Update $\boldsymbol{\epsilon}$.} Given $\mathbf{w}$, the subproblem on $\boldsymbol{\epsilon}_{i(\mathcal{Y}_t)}$ becomes
\begin{equation} \label{eq:8}
\begin{aligned}
&\min_{\boldsymbol{\epsilon}_i}\|\boldsymbol{\epsilon}_{i(\mathcal{Y}_t)}+J_{i(\mathcal{Y}_t)}-(S_{i(\mathbf{y}_{i(0)})}-S_{i(\mathcal{Y}_t)})\|_2^2\\
&\mathrm{s.t.}\;\boldsymbol{\epsilon}_{i(\mathcal{Y}_t)}\geq0
\end{aligned}
\end{equation}
Therefore, this subproblem has a closed-form solution,
\begin{equation} \label{eq:9}
\boldsymbol{\epsilon}_{i(\mathcal{Y}_t)}=\max\{0,S_{i(\mathbf{y}_{i(0)})}-S_{i(\mathcal{Y}_t)}-J_{i(\mathcal{Y}_t)}\}
\end{equation}

\indent \textbf{Update $\mathbf{w}$.} Given $\boldsymbol{\epsilon}_i$, the subproblem on $\mathbf{w}$ becomes
\begin{equation} \label{eq:10}
\min_{\mathbf{w}}\sum_{d=1}^D\|\boldsymbol{\gamma}\scriptstyle\bigodot\textstyle\mathbf{w}^d\|^2+C\|S_{i(\mathcal{Y}_t)}-(S_{i(\mathbf{y}_{i(0)})}-J_{i(\mathcal{Y}_t)}-\boldsymbol{\epsilon}_{i(\mathcal{Y}_t)})
\|^2_2
\end{equation}

\indent In order to obtain a simple expression of the normal equations, $\boldsymbol{\rho}_{i(\mathcal{Y}_t)}=S_{i(\mathbf{y}_{i(0)})}-J_{i(\mathcal{Y}_t)}-\boldsymbol{\epsilon}_{i(\mathcal{Y}_t)}$ is defined as the confidence label for each training sample. Then, using Parseval's formula, Eq.~\ref{eq:10} can be transformed to the Fourier frequency domain as,
\begin{equation} \label{eq:11}
\min_{\mathbf{w}}\sum_{d=1}^D\|\hat{\boldsymbol{\gamma}}\scriptstyle\bigotimes\textstyle\hat{\mathbf{w}}^d\|^2+
C\|\hat{S}_{i(\mathcal{Y}_t)}-\hat{\boldsymbol{\rho}}_{i(\mathcal{Y}_t)}\|^2_2
\end{equation}
where the Fourier coefficients of confidence labels and penalty function are $\hat{\boldsymbol{\rho}}_{i(\mathcal{Y}_t)}[k]=\hat{S}_{i(\mathbf{y}_{i(0)})}[k]-\hat{J}_{i(\mathcal{Y}_t)}[k]-\hat{\boldsymbol{\epsilon}}_{i(\mathcal{Y}_t)}[k]$ and $\hat{\boldsymbol{\gamma}}[k]=\frac{1}{T}\int_0^T\boldsymbol{\gamma}(t)e^{-i\frac{2\pi k}{T}t}$ respectively, and $\hat{J}_{i(\mathcal{Y}_t)}[k]$ can be straightforwardly deduced from $\hat{m_{\mathbf{x}}}[k]=\frac{\sqrt{2\pi\sigma^2}}{T}\exp\Big(-2\sigma^2(\frac{\pi k}{T})^2-i\frac{2\pi k}{T}\mathbf{y}_{i(0)}\Big)$.

\indent Finally, subproblem on $\mathbf{w}$ can be addressed by the following normal equation
\begin{equation} \label{eq:12}
\Big(\sum_{d=1}^D(\hat{\Phi}_{i(0)}^d)^H\hat{\Phi}_{i(0)}^d+\frac{1}{C}(\hat{\boldsymbol{\gamma}})^H\hat{\boldsymbol{\gamma}}\Big)\hat{\mathbf{w}}^d
=(\hat{\Phi}_{i(0)}^d)^H(\hat{\boldsymbol{\rho}}_{i(\mathcal{Y}_t)})^H
\end{equation}
Here, $\hat{\Phi}_{i(0)}^d=\hat{\Phi}^d\Big\{\varphi(\mathbf{x}_i,\mathbf{y}_{i(0)})\Big\}$, and $\mathbf{a}^H$ represents the complex conjugation of a complex matrix $\mathbf{a}$. In our approach, we apply the CG method mentioned in C-COT~\cite{ccot} to solve Eq.~\ref{eq:12} iteratively, since it can utilize the sparsity structure effectively.

\indent Similar to the original DCF framework, we also derive a training equation that does not exploit the penalty function $\boldsymbol{\gamma}$. For computational efficiency, we set $\boldsymbol{\gamma}=\mathbf{I}$, where $\mathbf{I}$ is an identity matrix. The optimization problem Eq.~\ref{eq:11} can be rewritten as,
\begin{equation} \label{eq:13}
\min_{\mathbf{w}}\sum_{d=1}^D\|\hat{\mathbf{w}}^d\|^2+C\|\hat{S}_{i(\mathcal{Y}_t)}-\hat{\boldsymbol{\rho}}_{i(\mathcal{Y}_t)}\|^2_2
\end{equation}
\indent In this case, Eq.~\ref{eq:13} is similar to conventional DCF-based trackers which aim to find optimal correlation filters by minimizing a mean squared error between the estimated confidence score map and the desired output. The normal equation Eq.~\ref{eq:13} of $\mathbf{w}$ can be expressed as,
\begin{equation} \label{eq:14}
\Big(\sum_{d=1}^D(\hat{\Phi}_{i(0)}^d)^H\hat{\Phi}_{i(0)}^d+\frac{1}{C}\mathbf{I}\Big)\hat{\mathbf{w}}^d=
(\hat{\Phi}_{i(0)}^d)^H(\hat{\boldsymbol{\rho}}_{i(\mathcal{Y}_t)})^H
\end{equation}
$\mathbf{w}$ can be obtained by preforming element-wise operations in the Fourier frequency domain,
\begin{equation} \label{eq:15}
\hat{\mathbf{w}}^d=\frac{(\hat{\Phi}_{i(0)}^d)^H(\hat{\boldsymbol{\rho}}_{i(\mathcal{Y}_t)})^H}
{\sum_{d=1}^D(\hat{\Phi}_{i(0)}^d)^H\hat{\Phi}_{i(0)}^d+\frac{1}{C}\mathbf{I}}
\end{equation}
where the $\div$ symbol denotes the element-wise division. We can easily recover structural correlation filters $\mathbf{w}$ using Fourier transforms in the continuous spatial domain. Compared with canonical DCF-based trackers, our collaborative optimization performs convolution in the continuous-domain by exploiting interpolated feature maps. Further, the proposed approach can create confidence score maps of the target by solving the continuous structured function directly, which is more discriminative than ridge regression models~\cite{mosse,kcf}.
\subsection{Ensemble post-processor}\label{sec:appro3}
To detect the target in a new frame $(i+1)$, all circular shifts of the possible target image patch $\mathbf{x}_{i+1}$ of the previous size $M_i\times N_i$ centered around the previous target position $\mathbf{y}_{i}$ are considered as searching samples. This is performed using the same procedure as for training samples in section~\ref{sec:appro1}.

\indent We define searching samples as circular input-output pairs $(\mathcal{X}_s,\mathcal{Y}_s)_{i+1}$, hence, the input space $\mathcal{X}_s$ denotes the possible target image patches, and the output space $\mathcal{Y}_s = \Big\{\mathbf{y}_{i(p)}\,|\,p\in\{0,\ldots,\Lambda-1\}\Big\}$ is now indicating the possible target positions on the circular shifted image patches. Since circular and structural operators are updated by iterative training structural correlation filters in the Fourier frequency domain, the primal confidence score map $S$ is obtained as,
\begin{equation} \label{eq:31}
\begin{aligned}
S=\mathcal{F}^{-1}\Big(\sum_{d=1}^D\mathcal{P}_T(\hat{S}_{i(\mathcal{Y}_s)})\Big)
\end{aligned}
\end{equation}
Notice that before performing IDFT, $\Phi^d\Big\{\varphi(\mathbf{x}_{i},\mathbf{y}_{i(0)})\Big\}$ has been interpolated in the continuous spatial domain, we use a padding operator $\mathcal{P}_T$ to zero-padding the high frequencies of $\hat{S}_{i(\mathcal{Y}_s)}[k]=\sum_{d=1}^D\hat{\textbf{w}}^d[k]\hat{\Phi}^d\Big\{\varphi(\mathcal{X}_s,\mathcal{Y}_s)_{i+1}\Big\}[k]$, hence, $S$ has a constant size of $T$.

However, primal confidence score maps obtained from only a single-layer deep feature map may sometimes be too weak to deal with challenging scenarios. Robust trackers should imperatively have the better diversity of features. We propose an effective ensemble post-processor that leverages multi-layer deep feature maps extracted from multiple deep networks for visual tracking. The ensemble post-processor can substantially improve tracking performance by coalescing primal confidence score maps based on relative entropy.

\indent For primal confidence score map set $S=\{S_1,S_2,...,S_L\}$ based on $L$-layer deep feature maps, the primal confidence score map $S_l\in S$ consists of a probability distribution $S_{l,\mathbf{y}_{i(p)}}$, which also can be considered as a probability map. The probability distribution is subjected to $\sum_{\mathbf{y}_{i(p)}\in\mathcal{Y}_s} S_{l,\mathbf{y}_{i(p)}}=1$ and presents the possibility that the possible position $\mathbf{y}_{i(p)}$ becomes the estimated target position. Further, in order to find the optimal confidence score map $R$, we can minimize the relative entropy, i.e., Kullback-Leibler divergence, between the primal confidence score map $S_l$ of each feature layer and the optimal confidence score map $R$. Relative entropy can be calculated as,
\begin{equation} \label{eq:16}
KL(S_l\|R)_{i+1}=\sum_{\mathbf{y}_{i(p)}\in\mathcal{Y}_s} S_{l,\mathbf{y}_{i(p)}}\log{\frac{S_{l,\mathbf{y}_{i(p)}}}{R_{\mathbf{y}_{i(p)}}}}
\end{equation}
Then, we can obtain the optimal confidence score map $R$ by
\begin{equation} \label{eq:17}
\begin{aligned}
&\arg\min_R\sum_{l=1}^{L}KL(S_l\|R)_{i+1}\\
&s.t. \sum_{\mathbf{y}_{i(p)}\in\mathcal{Y}_s} R_{\mathbf{y}_{i(p)}}=1
\end{aligned}
\end{equation}
where $S_{l,\mathbf{y}_{i(p)}}$ indicates the probability associated with the searching target position ${\mathbf{y}_{i(p)}}$ on the $l$-th layer primal confidence score map $S_l$ and $R_{\mathbf{y}_{i(p)}}$ indicates the probability on the optimal confidence score map $R$.

\indent Nonetheless, deep features extracted from pre-trained networks always contain various noise, so that primal confidence score maps obtained using Eq.~\ref{eq:5} also have much noise. In order to filter these noise and obtain a more reliable confidence score map, we propose an efficient collaborative filtering operation. Similar to MFCMT~\cite{mfcmt}, since all primal confidence score maps have the same resolution because of exploiting interpolation, we can directly weight a primal confidence score map using all other primal confidence score maps. Therefore, the confidence score map created by fusing any two different layers is more reliable and accurate. The collaborative filter can be formulated as
\begin{equation} \label{eq:18}
S_{m,n}=S_mS_n
\end{equation}
where $m\in\{1,2,\ldots,L-1\}$ and $n\in\{m+1,m+2,\ldots,L\}$. After filtering $L$ primal confidence score maps, we get $\frac{L(L-1)}{2}$ weighted confidence score maps $\Theta=\{S_{1,2},S_{1,3},\ldots,S_{1,L},S_{2,3},\ldots,S_{L-1,L}\}$ with less noise. With this collaborative filter, if two different primal confidence score maps have similar confidence score at the same position, the weighted confidence score at that position will be higher, while at other positions will have lower weighted confidence scores. Using these weighted confidence score maps, Eq.~\ref{eq:17} can be reformulated as
\begin{equation} \label{eq:19}
\begin{aligned}
&\arg\min_R\sum_{S_l\in\Theta}\sum_{\mathbf{y}_{i(p)}\in\mathcal{Y}_s} S_{l,\mathbf{y}_{i(p)}}\log{\frac{S_{l,\mathbf{y}_{i(p)}}}{R_{\mathbf{y}_{i(p)}}}}\\
&s.t. \sum_{\mathbf{y}_{i(p)}\in\mathcal{Y}_s} R_{\mathbf{y}_{i(p)}}=1
\end{aligned}
\end{equation}
\indent We exploit the Lagrange multiplier method~\cite{mcft} to solve this problem. The optimal confidence score map can be calculated as
\begin{equation} \label{eq:20}
R=\frac{2}{L(L-1)}\sum_{S_l\in\Theta}S_l
\end{equation}

\indent Interestingly, although the optimal confidence score map $R$ is obtained by averaging all weighted confidence score maps, it enhances the robustness of the tracking result. Finally, the target position $\mathbf{y}_{i+1}$ is determined by maximizing the optimal confidence score map $R$.

\indent In a deep appearance feature network, the shallow layer contains high-resolution and low-level spatial information, while the deep layer involves low-resolution and high-level semantic information. In a deep motion feature network, the deep layer encodes high-level motion information at coarse resolution. Therefore, we can get more discriminative confidence score maps by combining these multiple layers deep features. In Section~\ref{sec:exp}, we will show that our approach outperforms several state-of-the-art trackers both in terms of accuracy and efficiency.

\indent Some works~\cite{dsst,samf} demonstrate that effective scale estimation methods are indispensable. In order to improve the accuracy of our proposed approach, we construct a rectangular image patch pyramid by exploiting scale adaptation scheme proposed in CFWCR~\cite{cfwcr}. The pyramid is centered at the estimated target position in the previous frame. For scale factors $\Big\{a^\tau\,|\,\tau=\lfloor-\frac{\Delta-1}{2}\rfloor,\ldots,\lfloor\frac{\Delta-1}{2}\rfloor\Big\}$, where $\Delta$ denotes the number of scale layers, we can obtain a set of input-output pairs $(\mathbf{x}^\delta,\mathbf{y}_{p}^\delta)$ of size $a^\tau M\times a^\tau N$ on the $\delta$-th scaled input image. After that, we extract multiple interpolated deep feature maps according to Eq.~\ref{eq:4}. Structural correlation filters $\mathbf{w}$ trained from the previous frame can be applied to each deep feature map $\Phi(\mathbf{x}^\delta,\mathbf{y}_{p}^\delta)$ to get the primal confidence score map. Next, according to Eq.~\ref{eq:20}, we can get an optimal confidence score map $R^\delta$ of each scaled layer $\delta$. Finally, the estimated target position $\mathbf{y}_{i}$ of the current frame $i$ is determined by the peak value over all optimal confidence score maps $R^\Delta$.
\begin{algorithm}[t]
\renewcommand{\algorithmicrequire}{\textbf{Input:}}
\renewcommand{\algorithmicensure}{\textbf{Output:}}
\caption{The proposed approach}
\label{alg:1}
\begin{algorithmic}[1]
\Require image sequence $\{\mathbf{I}_i\}_{i=1}^I$, initial target position $\mathbf{y}_1$ and size $M_1\times N_1$, scale layers $\Delta$, feature layers $L$
\Ensure estimated target position $\{\mathbf{y}_i\}_{i=2}^I$ and size $\{M_i\times N_i\}_{i=2}^I$
\State Initialize structural correlation filters $\{\mathbf{w}_{0,l}\}_{l=1}^L=0$,
\State $\vartriangleright$ Update structural correlation filters using 25 collaborative optimization iterations in the first frame
\State Crop training samples $\mathbf{x}_1$ of size $M_1\times N_1$ from $\mathbf{I}_1$ at the initial target position $\mathbf{y}_1$,
\State Extract $L$ layers continuous-domain deep feature maps $\{\Phi_{1,l}\}_{l=1}^L$ of searching samples $\mathbf{x}_1$ using Eq.~\ref{eq:4},
\For{$j=1,\ldots,25$}
\State Update $\boldsymbol{\epsilon}_j$ with $\mathbf{w}_{j-1}$ using Eq.~\ref{eq:9},
\State Update structural correlation filters $\mathbf{w}_j$ with $\mathbf{w}_{j-1}$, $\Phi_1$, $\boldsymbol{\epsilon}_j$ and $\mathbf{y}_1$ using Eq.~\ref{eq:5}, Eq.~\ref{eq:9} and Eq.~\ref{eq:12},
\EndFor
\State $\mathbf{w}_0 \leftarrow \mathbf{w}_{25}$,
\For{$i=2,\ldots,T$}
\State $\vartriangleright$ Estimate the target position and size from $\Delta$ different scaled searching samples
\For{$\delta=1,\ldots,\Delta$}
\State Crop searching samples $\mathbf{x}_i$ of size $M_{i-1}^\delta\times N_{i-1}^\delta$ from $\mathbf{I}_i$ at the previously estimated target position $\mathbf{y}_{i-1}$,
\State Extract $L$ layers continuous-domain deep feature maps $\{\Phi_{i,l}\}_{l=1}^L$ of searching samples $\mathbf{x}_i$ using Eq.~\ref{eq:4},
\For{$l=1,\ldots,L$}
\State Calculate the primal confidence score map $S_l$ of each layer continuous-domain deep features map $\Phi_{i,l}$ with $\mathbf{w}_{0,l}$ using circular and structural operator $S_\mathbf{w}$ as Eq.~\ref{eq:5},
\EndFor
\State Calculate the scaled optimal confidence score map $R^\delta$ using Eq.~\ref{eq:18} and Eq.~\ref{eq:20},
\EndFor
\State Set $\mathbf{y}_i$ to the position corresponding to the maximum score over all scaled optimal confidence score map $R^\Delta$
\State $M_t\times N_t \leftarrow M_{t-1}^\delta\times N_{t-1}^\delta$, where $R^\delta$ contains the maximum confidence score,
\State Crop training samples $\mathbf{x}_i$ of size $M_i\times N_i$ from $\mathbf{I}_i$ at the current estimated target position $\mathbf{y}_{i}$,
\State Extract $L$ layers continuous-domain deep features maps $\{\Phi_{i,l}\}_{l=1}^L$ of the search region $\mathbf{x}_i$ using Eq.~\ref{eq:4},
\State $\vartriangleright$ Update structural correlation filters using three collaborative optimization iterations
\For{$j=1,\ldots,3$}
\State Update $\boldsymbol{\epsilon}_j$ with $\mathbf{w}_{j-1}$ using Eq.~\ref{eq:9},
\State Update structural correlation filters $\mathbf{w}_j$ with $\mathbf{w}_{j-1}$, $\Phi_i$, $\boldsymbol{\epsilon}_j$ and $\mathbf{y}_i$ using Eq.~\ref{eq:5}, Eq.~\ref{eq:9} and Eq.~\ref{eq:12},
\EndFor
\State $\mathbf{w}_0 \leftarrow \mathbf{w}_{3}$,
\EndFor
\end{algorithmic}
\end{algorithm}

\indent For feature maps with $M\times N\times D$ dimensions, we should solve $MN$ subproblems, and the complexity of each subproblem is a linear equation with $D$ variables. Due to the computational efficiency of CG method, each subproblem can be solved in $\mathcal{O}(D)$. Since the optimizing process requires DFT, IDFT and element-wise operations in each iteration, the complexity of solving $\mathbf{w}$ is $\mathcal{O}\Big(DMN \log (MN)\Big)$. The $\boldsymbol{\epsilon}$ update problem can be solved in element-wise, which has the cost of $\mathcal{O}(MN)$. Thus, the overall complexity to compute the online collaborative optimization algorithm is $\mathcal{O}\Big(DMNI_{CG}\log(MN)\Big)$, where $I_{CG}$ indicates the number of CG iterations. In a word, our approach is especially suitable for high-dimensional features and dense training samples. We summarize our proposed approach in Algorithm~\ref{alg:1}.
\subsection{Nonlinear extension}
It is well known that the kernel trick can improve tracking accuracy further by allowing detection on richer high-dimensional features~\cite{kcf,scf}. By exploiting the kernel function $ K_{mn}=\kappa\bigg\langle\Phi\Big\{\varphi(\mathbf{x}_i,\mathbf{y}_{i(m)})\Big\},\Phi\Big\{\varphi(\mathbf{x}_i,\mathbf{y}_{i(n)})\Big\}\bigg\rangle$, e.g., a Gaussian RBF kernel $\kappa\bigg\langle\Phi\Big\{\varphi(\mathbf{x}_i,\mathbf{y}_{i(m)})\Big\},\Phi\Big\{\varphi(\mathbf{x}_i,\mathbf{y}_{i(n)})\Big\}\bigg\rangle=
\exp\Big(-\frac{1}{\sigma_k^2}\Big\|\Phi\Big\{\varphi(\mathbf{x}_i,\mathbf{y}_{i(m)})\Big\}-\Phi\Big\{\varphi(\mathbf{x}_i,\mathbf{y}_{i(n)})\Big\}\Big\|^2\Big)$, the proposed CSOT can be easily posed to the implicit nonlinear space~\cite{kcf}. Structural correlation filters are determined as the weighted sum of circular training samples $\mathbf{w}=\displaystyle\sum_{\mathbf{y}_{i(p)}\in\mathcal{Y}_t}\boldsymbol{\alpha}_{\mathbf{y}_{i(p)}}\Phi\Big\{\varphi(\mathbf{x}_i,\mathbf{y}_{i(p)})\Big\}$, where $\mathbf{y}_{i(0)}$ is the estimated target position at frame $i$ and $\boldsymbol{\alpha}$ is the parameter vector to be learned. Hence, we have
\begin{equation} \label{eq:22}
\begin{aligned}
\|\mathbf{w}\|^2
&=(\boldsymbol{\alpha})^T\mathbf{K}_i\boldsymbol{\alpha}\\
&=(\boldsymbol{\alpha})^T\mathcal{F}^{-1}(\hat{k}_{\Phi_{i}^{(0)}\Phi_{i}^{(0)}}\scriptstyle\bigodot\textstyle\hat{\boldsymbol{\alpha}})
\end{aligned}
\end{equation}
where $\Phi_{i}^{(0)}=\Phi\Big\{\varphi(\mathbf{x}_i,\mathbf{y}_{i(0)})\Big\}$, $\mathbf{K}_i$ denotes a circular kernel matrix whose elements are $K_{mn}$, and $k_{\Phi_{i}^{(0)}\Phi_{i}^{(0)}}$ which obtained by the kernel auto-correlation of $\Phi_{i}^{(0)}$ is the first row of $\mathbf{K}_i$.

\indent Based on Eq.~\ref{eq:10} and Eq.~\ref{eq:22}, The optimization of kernelized CSOT is formulated as,
\begin{equation} \label{eq:23}
\begin{aligned}
&\min_{\boldsymbol{\alpha}}\boldsymbol{\alpha}^T\mathbf{K}_i\boldsymbol{\alpha}\|{\boldsymbol{\gamma}}\|_2^2+
C\|S_{i(\mathcal{Y}_t)}-(S_{i(\mathbf{y}_{i(0)})}-J_{i(\mathcal{Y}_t)}-\boldsymbol{\epsilon}_{i(\mathcal{Y}_t)})\|^2_2\\
&\mathrm{s.t.}\;\boldsymbol{\epsilon}_{i(\mathcal{Y}_t)}\geq0
\end{aligned}
\end{equation}
where $S_{i(\mathcal{Y}_t)}=\Big(\Phi\Big\{\boldsymbol{\varphi}(\mathcal{X}_t,\mathcal{Y}_t)_i\Big\}\Big)^T\mathbf{w}=\mathbf{K}_i\boldsymbol{\alpha}$ is the primal confidence score map.

\indent Similar to Eq.~\ref{eq:11}, $\boldsymbol{\rho}_{i(\mathcal{Y}_t)}$ is plugged into Eq.~\ref{eq:23}
\begin{equation} \label{eq:24}
\min_{\boldsymbol{\alpha}}\boldsymbol{\alpha}^T\mathcal{F}^{-1}(
\hat{k}_{\Phi_{i}^{(0)}\Phi_{i}^{(0)}}\scriptstyle\bigodot\textstyle\hat{\boldsymbol{\alpha}})\|{\boldsymbol{\gamma}}\|_2^2+
C\|\mathcal{F}^{-1}(
\hat{k}_{\Phi_{i}^{(0)}\Phi_{i}^{(0)}}\scriptstyle\bigodot\textstyle\hat{\boldsymbol{\alpha}})-\hat{\boldsymbol{\rho}}_{i(\mathcal{Y}_t)}\|^2_2
\end{equation}

\indent Then, CG method is employed to iteratively solve $\boldsymbol{\alpha}$ in the Fourier frequency domain as,
\begin{equation} \label{eq:25}
\hat{\boldsymbol{\alpha}}=\Big(\hat{k}_{\Phi_{i}^{(0)}\Phi_{i}^{(0)}}+\frac{1}{C}(\hat{\boldsymbol{\gamma}})^H\hat{\boldsymbol{\gamma}}\Big)^{-1}(\hat{\boldsymbol{\rho}}_{i(\mathcal{Y}_t)})^H
\end{equation}

\indent The closed-form solution to optimize $\boldsymbol{\alpha}$ also can be expressed without exploiting the spatial penalty function $\boldsymbol{\gamma}$ similar to Eq.~\ref{eq:13},
\begin{equation} \label{eq:27}
\hat{\boldsymbol{\alpha}}=\frac{\hat{\boldsymbol{\rho}}^H_{i(\mathcal{Y}_t)}}{\hat{k}_{\Phi_{i}^{(0)}\Phi_{i}^{(0)}}+\frac{1}{C}\mathbf{I}}
\end{equation}

\indent When a new frame $(i+1)$ comes out, after obtaining the searching samples $(\mathcal{X}_s,\mathcal{Y}_s)_{i+1}$, the primal confidence score maps for detection is
\begin{equation} \label{eq:27}
\begin{aligned}
S=\mathcal{F}^{-1}\Big(\mathcal{P}_T(\hat{k}_{\Phi_{i+1}^{(0)}\Phi_{i}^{(0)}}\scriptstyle\bigodot\textstyle\hat{\boldsymbol{\alpha}})\Big)
\end{aligned}
\end{equation}
where $\hat{k}_{\Phi_{i+1}^{(0)}\Phi_{i}^{(0)}}$ is the kernel circular correlation of $\Phi_{i+1}^{(0)}$ and $\Phi_{i}^{(0)}$, and $\Phi_{i+1}^{(0)}=\Phi\Big\{\varphi(\mathbf{x}_{i+1},\mathbf{y}_{i(0)})\Big\}$. Finally, similarly to section~\ref{sec:appro3}, we employ the ensemble post-processor to fuse all primal confidence score maps $S_l$, $l\in\{1,2,\ldots,L\}$, the target position is localized by searching the peak value of the optimal confidence score map $R$.
\section{Experiments} \label{sec:exp}
In this section, experimental evaluations are conducted to validate the effectiveness of the proposed CSOT. We first introduce the experimental setup and metrics. Then, we investigate the impact of the various components described in Section~\ref{sec:appro} by evaluating several variants of the proposed approach. Finally, our proposed approach is compared with several state-of-the-art trackers on the OTB and the VOT benchmarks.
\subsection{Experimental setup}
The proposed CSOT tracker is implemented in MATLAB R2014b using MatConvNet toolbox~\cite{matconvnet}~\footnote{\url{http://www.vlfeat.org/matconvnet}}. All experiments are performed on an Intel Core i5-4590 CPU @ 3.3GHz with 8GB RAM and a NVIDIA Tesla K80 GPU. In our experiments, we crop the searching sample to $5^2$ times the previously estimated target size and set the size of searching samples in a restricted area [200,\;300]. The cubic spline is used as the kernel to construct the interpolation function $b$ as described in the supplementary material of C-COT~\cite{ccot}. The regularization parameter $C$ in Eq.~\ref{eq:2} is set as 20000, and the spatial bandwidth $\sigma$ in the cost function Eq.~\ref{eq:3} is set as 0.1. Similar to SRDCF~\cite{srdcf}, the spatial regularizer $\boldsymbol{\gamma}$ in Eq.~\ref{eq:6} is constructed using a quadratic function $\boldsymbol{\gamma}(m,n)=0.1+3(m/M)^2+3(n/N)^2$, where 0.1 is the minimum value of structural correlation filters and 3 is the impact of the spatial regularizer. The kernel function in the nonlinear extension uses Gaussian Radial Basis Function with a spatial bandwidth of 0.2. Similar to CFWCR~\cite{cfwcr}, to conduct adaptive scale estimation, the number of scale layers $\Delta$ is set as 10 and the scale factor $a$ is set as 1.03. Online collaborative optimization is performed 25 iterations in the first frame. In the sequential frames, to converge to a desirable estimation of structural correlation filters, online collaborative optimization is iterated 3 times per frame. Each online collaborative optimization iteration involves two CG-iterations to learn structural correlation filters $\mathbf{w}$ using Eq.~\ref{eq:12}. And the CG is initialized with structural correlation filters learned in the previous frame. All parameters are fixed in the following experiments.
\subsection{Experimental metrics}
\indent All experiments are conducted on OTB50/2013/2015~\cite{otb2013,otb2015}~\footnote{\url{http://cvlab.hanyang.ac.kr/tracker_benchmark}} and VOT2017~\cite{vot2017}~\footnote{\url{http://www.votchallenge.net/challenges.html}} benchmarks. Two evaluation metrics are exploited on OTB benchmarks, i.e., distance precision (DP) and overlap precision (OP). DP is computed as the percentage of frames where the average Euclidean distance between the estimated target position and the ground truth is smaller than a preset threshold of 20 pixels. OP is the percentage of frames where the intersection-over-union~(IoU) exceeds a fixed threshold of 0.5, IoU is defined as $\frac{|\mathrm{B_t}\cap\mathrm{G_t}|}{|\mathrm{B_t}\cup\mathrm{G_t}|}$, where $B_t$ and $G_t$ are the estimated target bounding box and the ground truth, and $|\cdot|$ indicates the number of pixels in the overlap area. We use one-pass evaluation (OPE) that initialize the target position in the first frame and size, then run trackers throughout entire test sequences. Results are reported in precision and success plots. In success plots, trackers are ranked according to the area under the curve (AUC). We also report the speed of trackers by the average frames per second (FPS) over all sequences.

\indent For the VOT2017 benchmark, three evaluation metrics are exploited as the VOT committee suggested~\cite{vot}: (a) the accuracy, which measures the average overlap between the estimated bounding box and the ground truth, (b) the robustness, which measures the average failure times during tracking, (c) the expected average overlap (EAO), which measures the average overlap a tracker is expected to attain without re-initialization following a failure on a large collection of short-term sequences. Trackers are ranked according to EAO scores on the VOT2017 benchmark.
\subsection{Ablation experiments}
To verify the effectiveness of each component proposed in Section~\ref{sec:appro}, we implement and evaluate six variants of our approach on OTB50/2013/2015 benchmarks, which are respectively named as (a) CSOT-HC, (b) CSOT-KHC, (c) CSOT-DCNN, (d) CSOT-CNN, (e) CSOT-CNN2 and (f) CSOT.

\indent Characteristics of all variants are summarized in Table~\ref{table:3}. CSOT-HC without ensemble post-processor is implemented as the baseline, where handcrafted features such as HOG and CN are exploited. CSOT-KHC is developed using a Gaussian kernel with conventional handcrafted features. CSOT-CNN and CSOT-DCNN, without the ensemble post-processor, exploit only deep appearance features. CSOT-CNN2 is designed by incorporating deep motion features into CSOT-CNN. CSOT is implemented by integrating the ensemble post-processor into CSOT-CNN2. A noteworthy is CSOT, CSOT-CNN2, CSOT-CNN, CSOT-KHC and CSOT-HC employ both spatial regularization and implicit interpolation operators to incorporate continuous-domain features, while CSOT-DCNN exploits an explicit resampling strategy to extract features.
\begin{table}[t]
\centering\small
\caption{Characteristics of CSOT and its variants.}
\begin{tabular}{p{3.3cm}p{1.6cm}<{\centering}p{1.6cm}<{\centering}p{1.8cm}<{\centering}p{1.6cm}<{\centering}p{1.7cm}<{\centering}p{1.6cm}<{\centering}}
\toprule
Variants                    & CSOT-HC  & CSOT-KHC & CSOT-DCNN & CSOT-CNN  & CSOT-CNN2 & CSOT \\ \midrule
Deep appearance features    & $  -   $ & $  -   $ & $\surd $  & $\surd $  & $\surd $  & $\surd $   \\
Deep motion features        & $  -   $ & $  -   $ & $  -   $  & $  -   $  & $\surd $  & $\surd $   \\
Handcrafted features        & $\surd $ & $\surd $ & $  -   $  & $  -   $  & $  -   $  & $  -   $   \\
Multi-resolution features   & $\surd $ & $\surd $ & $  -   $  & $\surd $  & $\surd $  & $\surd $   \\
Ensemble post-processor     & $  -   $ & $  -   $ & $  -   $  & $  -   $  & $  -   $  & $\surd $   \\
Nonlinear kernel            & $  -   $ & $\surd $ & $  -   $  & $  -   $  & $  -   $  & $  -   $   \\\bottomrule
\end{tabular}
\label{table:3}
\end{table}
\begin{figure}[!tb]
\captionsetup{belowskip=0em}
\centering
\begin{minipage}{0.32\linewidth}\centerline{\includegraphics[width=\textwidth]{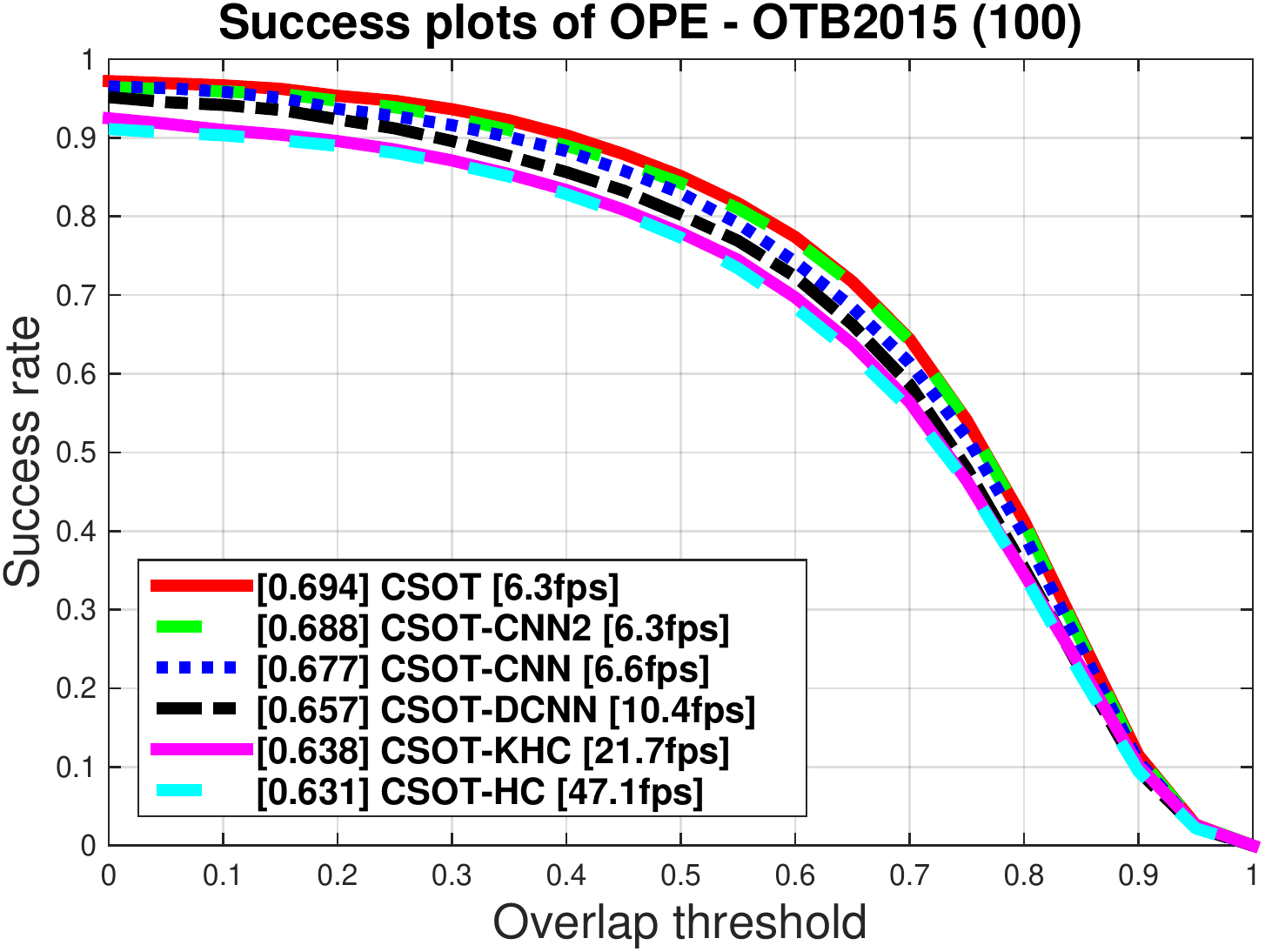}}\end{minipage}
\hfill\begin{minipage}{0.32\linewidth}\centerline{\includegraphics[width=\textwidth]{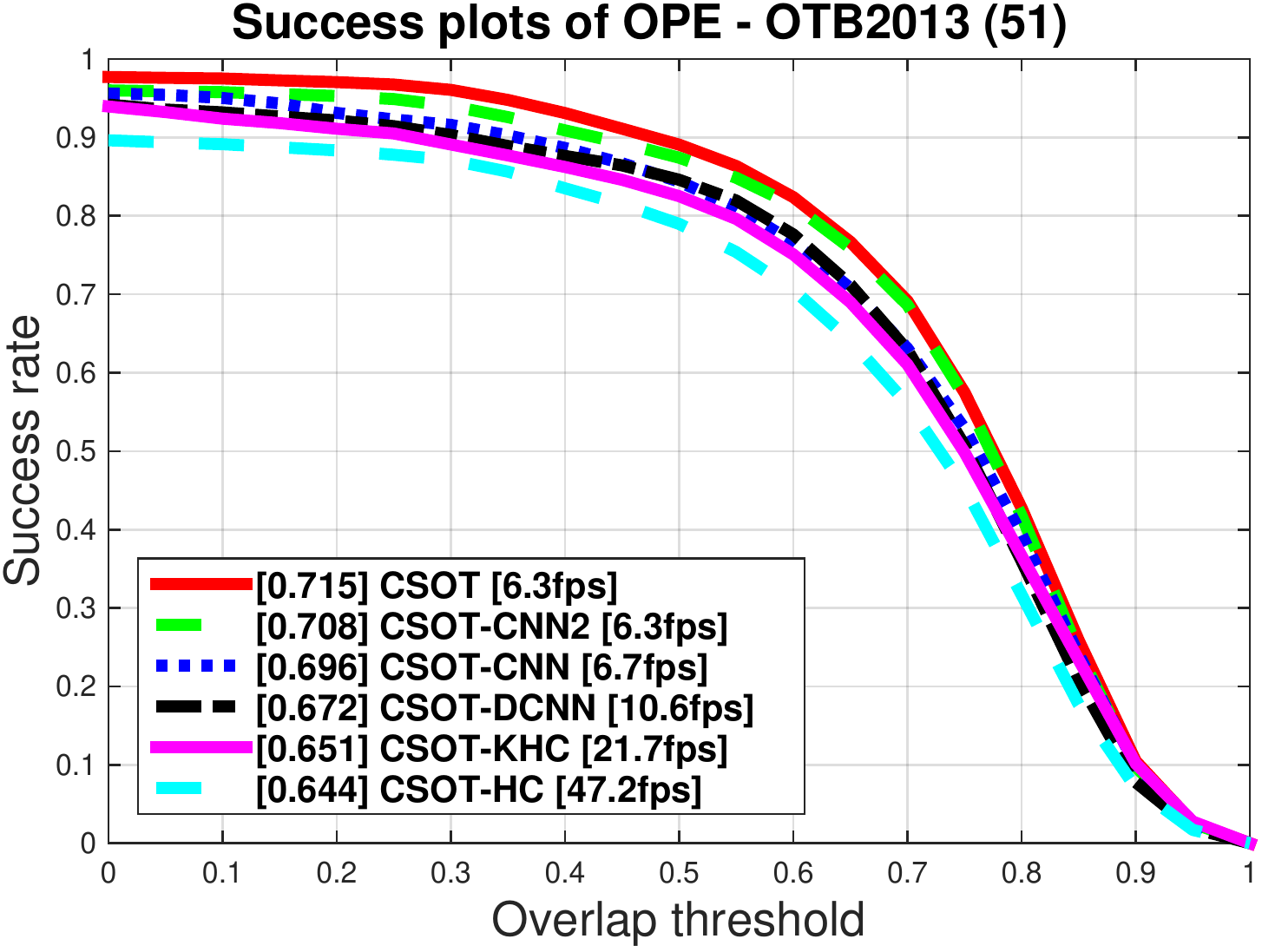}}\end{minipage}
\hfill\begin{minipage}{0.32\linewidth}\centerline{\includegraphics[width=\textwidth]{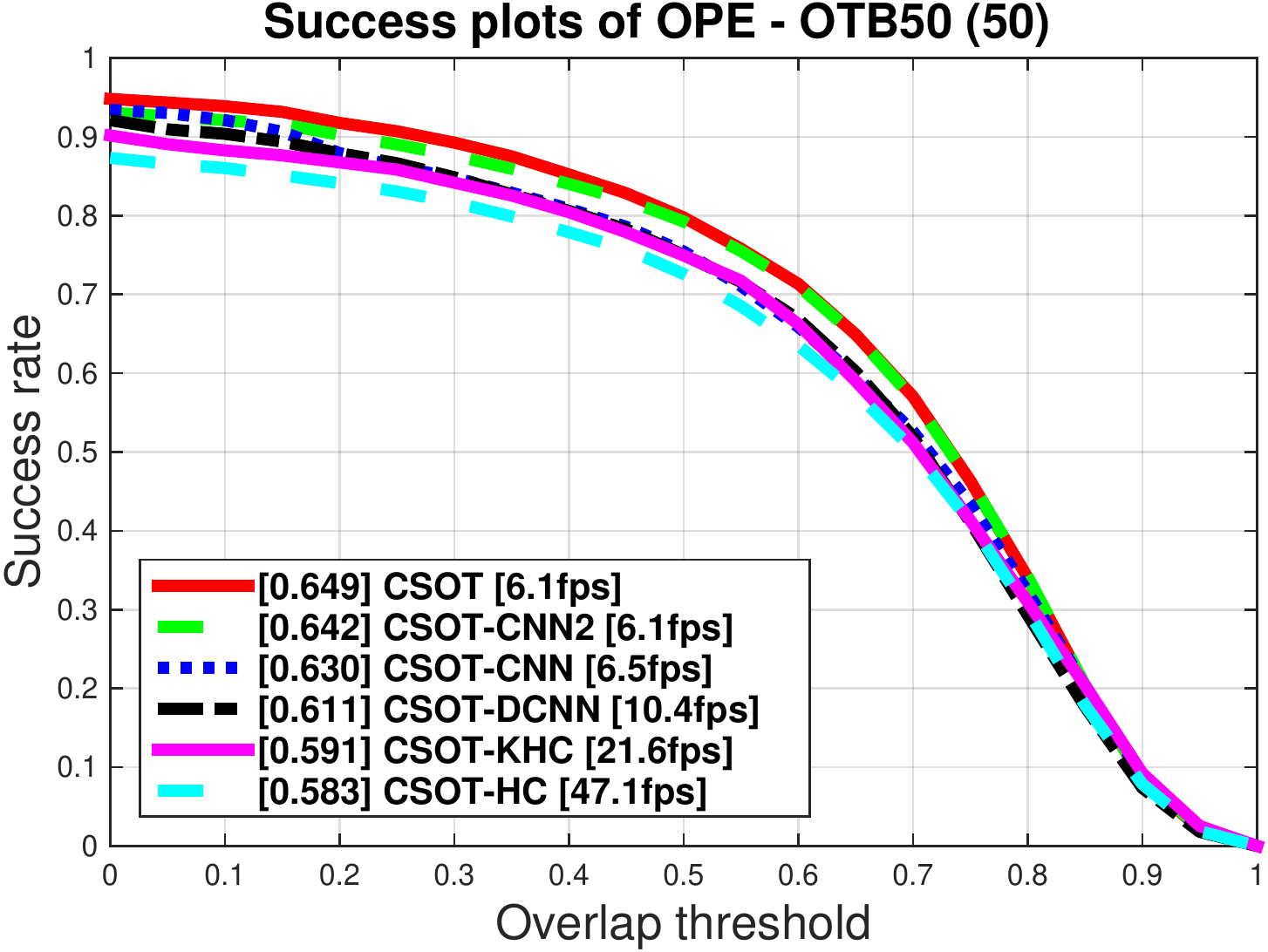}}\end{minipage}
\vspace{-1em}
\caption{Success plots of CSOT and its variants on OTB benchmarks~\cite{otb2013,otb2015}. The first value in the legend indicates the AUC score for each tracker. We also use the last number in the legend to present the mean speed of each tracker. Best viewed in color.}
\label{fig:3}
\end{figure}
\\\indent Success plots of CSOT and its variants on OTB2015~(with 100 sequences), OTB2013~(with 51 sequences) and OTB50~(with 50 sequences) benchmarks are illustrated in Fig.~\ref{fig:3}. Experimental results show that CSOT-HC gives AUC scores of 63.1\%, 64.4\% and 58.3\% on OTB2015, OTB2013 and OTB50 benchmarks, respectively. For the sake of clarity, the mean AUC score of these three OTB benchmarks is exploited as the evaluation metric in the following evaluations. Therefore, CSOT-HC obtains a mean AUC score of 61.9\%. Interestingly, CSOT-KHC obtains a mean AUC score of 62.7\%, the nonlinear extension is not helpful to improve results satisfactorily and reduces the tracking speed from 47.1 FPS to 21.6 FPS. Meanwhile, CSOT-CNN obtains a mean AUC score of 66.8\% which is 4.9\% higher than that of CSOT-HC. It demonstrates that deep appearance features can improve tracking performance obviously than conventional handcrafted features. We also employ CSOT-CNN to evaluate the efficiency of multi-resolution interpolation methods, it provides an absolute gain of 2.1\% compared with the discrete resampling version, i.e., CSOT-DCNN. By involving deep motion features to the CSOT-CNN, the mean AUC score of CSOT-CNN2 can be increased by 6.0\% compared to CSOT-HC. According to our analysis, it is apparent that involving deep motion features can boost tracking performance. The best result is provided by CSOT. Compared with CSOT-HC, CSOT achieves an absolute gain of 6.7\% in the mean AUC score. The results clearly demonstrate that deep appearance features and deep motion features are complementary, and lead to obtain the best results. Deep features extracted from only a single network have limited diversity, which is not sufficient to deal with various challenges. Moreover, CSOT adopts the ensemble post-processor to enhance diversity, which can enhance the performance to address problems of various challenging factors substantially.

\begin{figure}[!tb]
\captionsetup{belowskip=0em}
\centering
\begin{minipage}{0.49\linewidth}\centerline{\includegraphics[width=\textwidth]{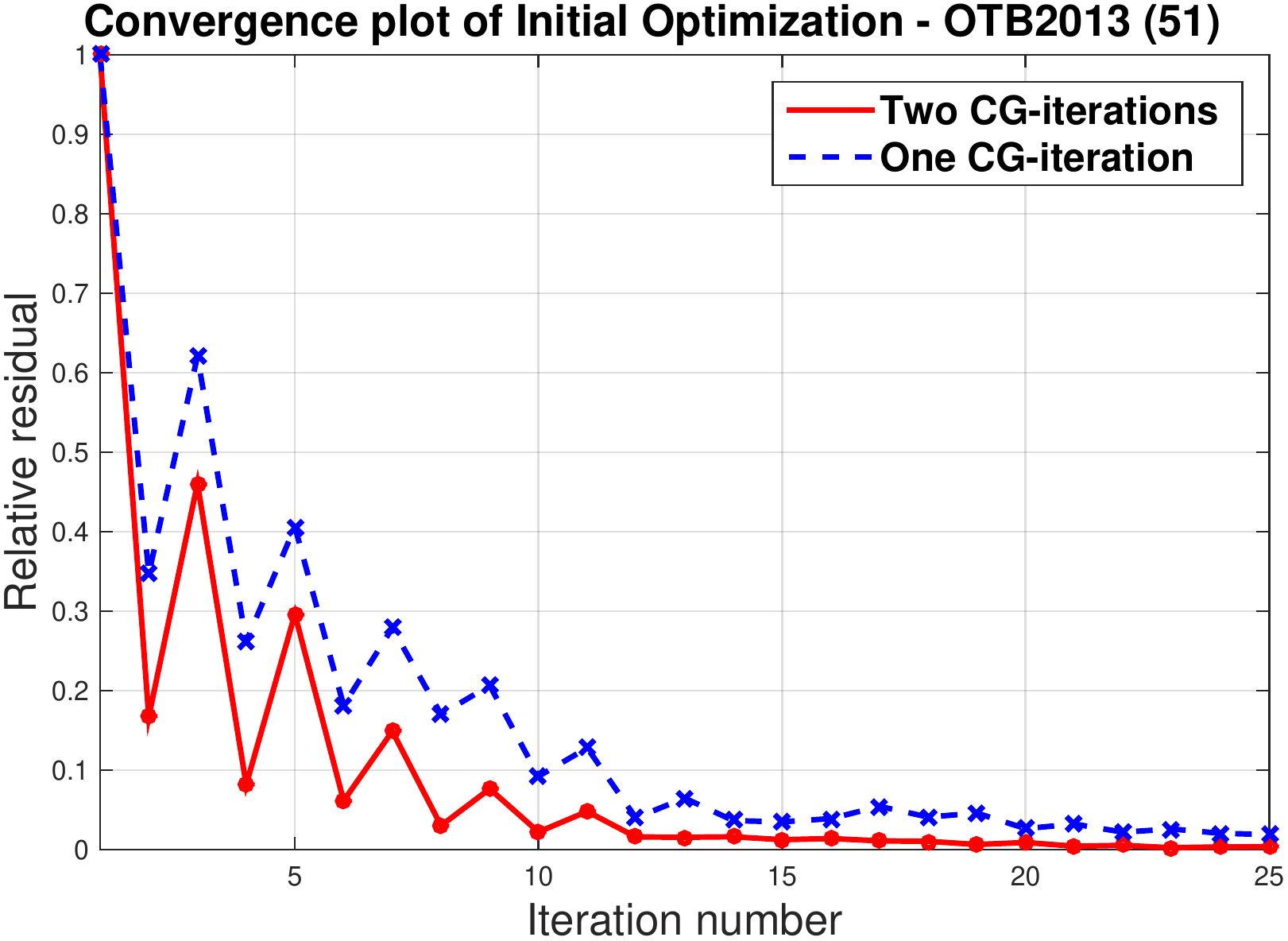}}\end{minipage}
\hfill\begin{minipage}{0.49\linewidth}\centerline{\includegraphics[width=\textwidth]{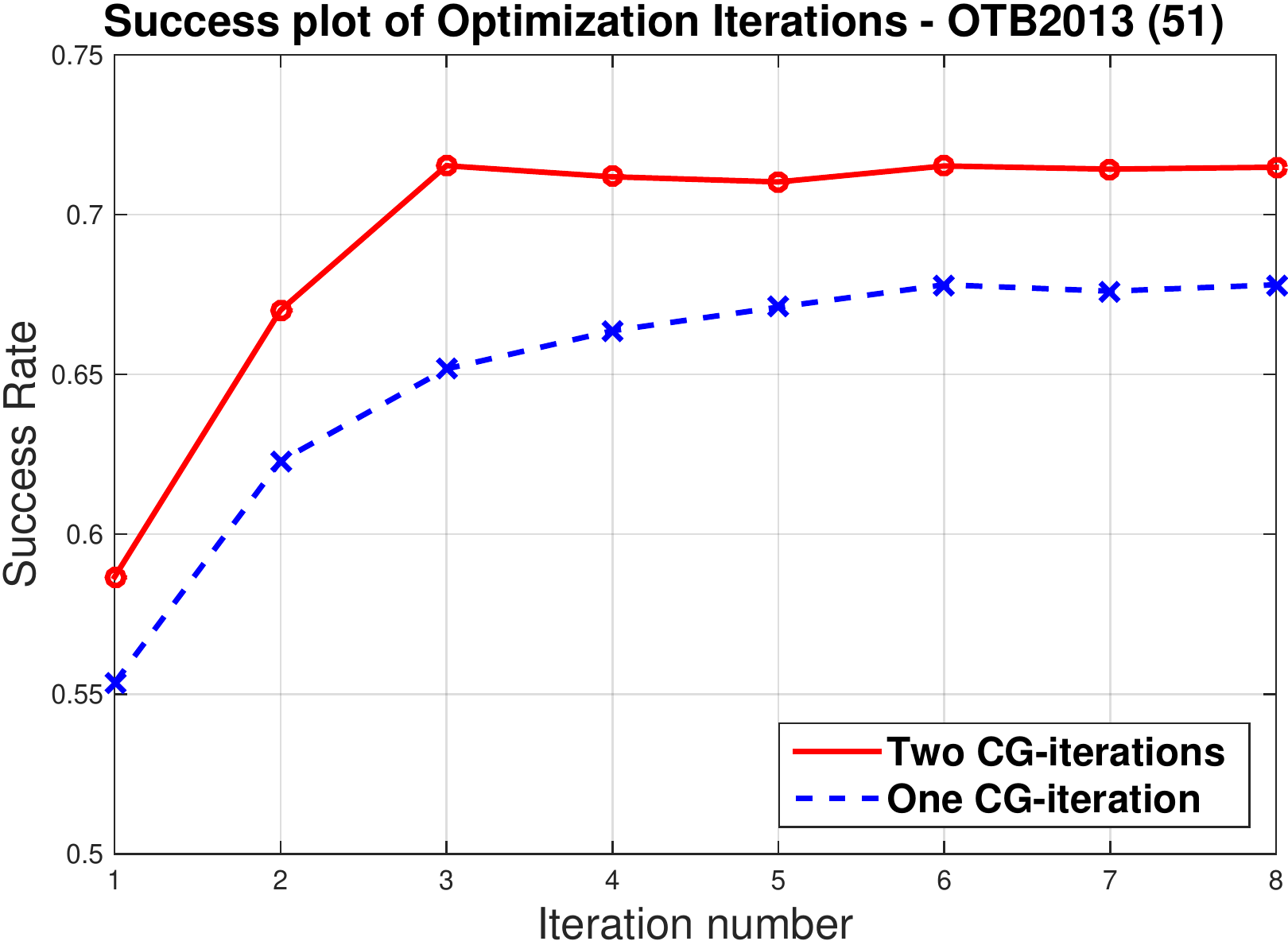}}\end{minipage}
%\vspace{-1em}
\caption{Convergence plot of initial online collaborative optimization (left) and success plot of different CG-iterations during online collaborative optimizations (right) on the OTB2013 benchmark.}
\label{fig:8}
\end{figure}

\indent For detailed analyses, we employ CSOT to evaluate the impact of the online collaborative optimization on the OTB2013 benchmark. Fig.~\ref{fig:8} shows the convergence plot of initial online collaborative optimization and the success plot of different CG-iterations during online collaborative optimizations. It can be observed that after 25 iterations of online collaborative optimization in the first frame, the relative residual falls close to zero, and this can lead to the desirable initial estimate of structural correlation filters. The success plot illustrates that after three iterations of the online collaborative optimization, where each optimization involves two CG-iterations, the AUC score tends to converge on the OTB2013 benchmark.

\indent Obviously, all experiment results demonstrate that those methods we proposed in Section~\ref{sec:appro} are reliable and effective. We will compare CSOT with many state-of-the-art tracking approaches in the following evaluations.
\subsection{Experiments on OTB}
\begin{figure}[t]
\captionsetup{belowskip=0em}
\centering
\begin{minipage}{0.32\linewidth}\centerline{\includegraphics[width=\textwidth]{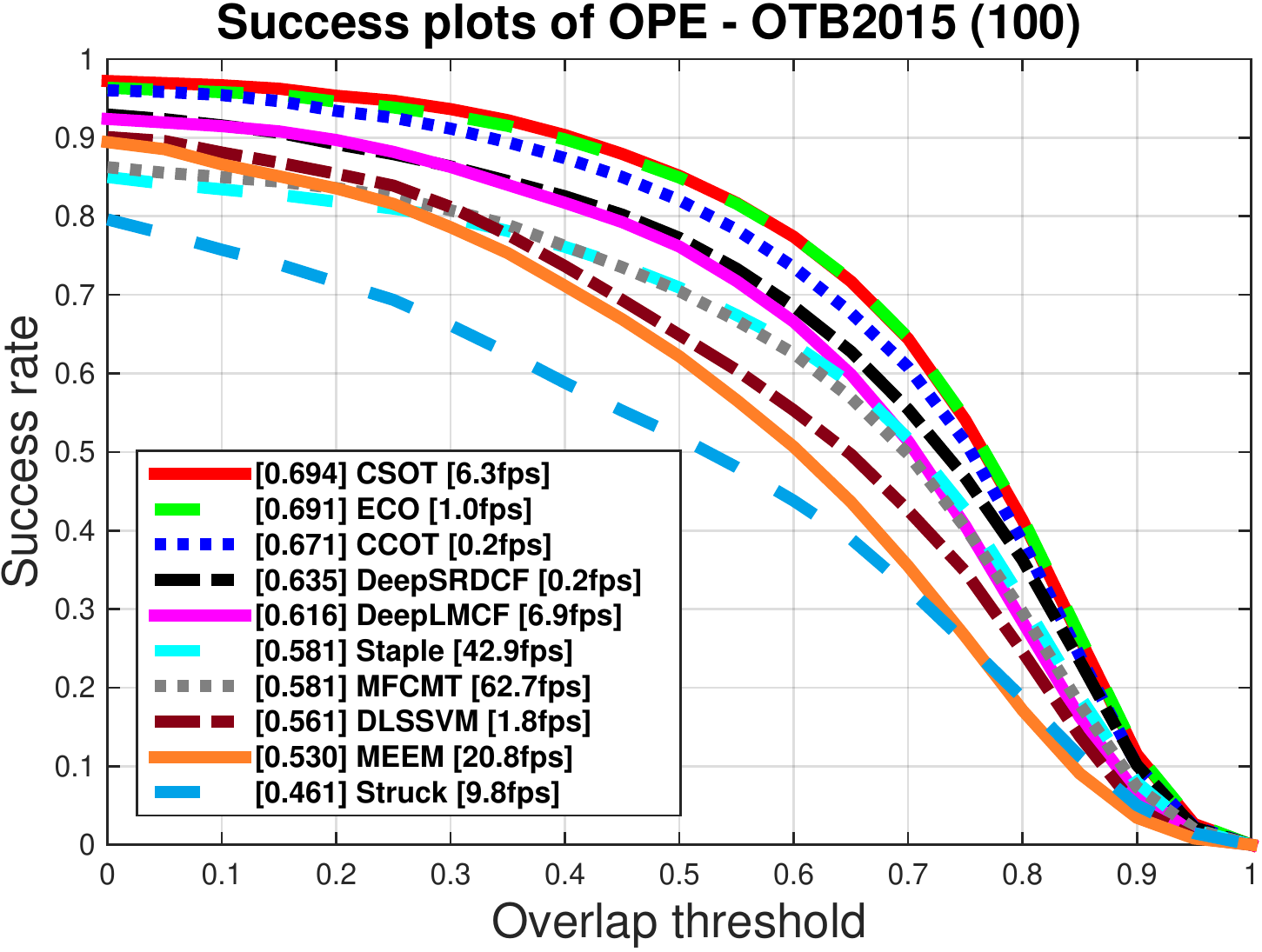}}\end{minipage}
\hfill\begin{minipage}{0.32\linewidth}\centerline{\includegraphics[width=\textwidth]{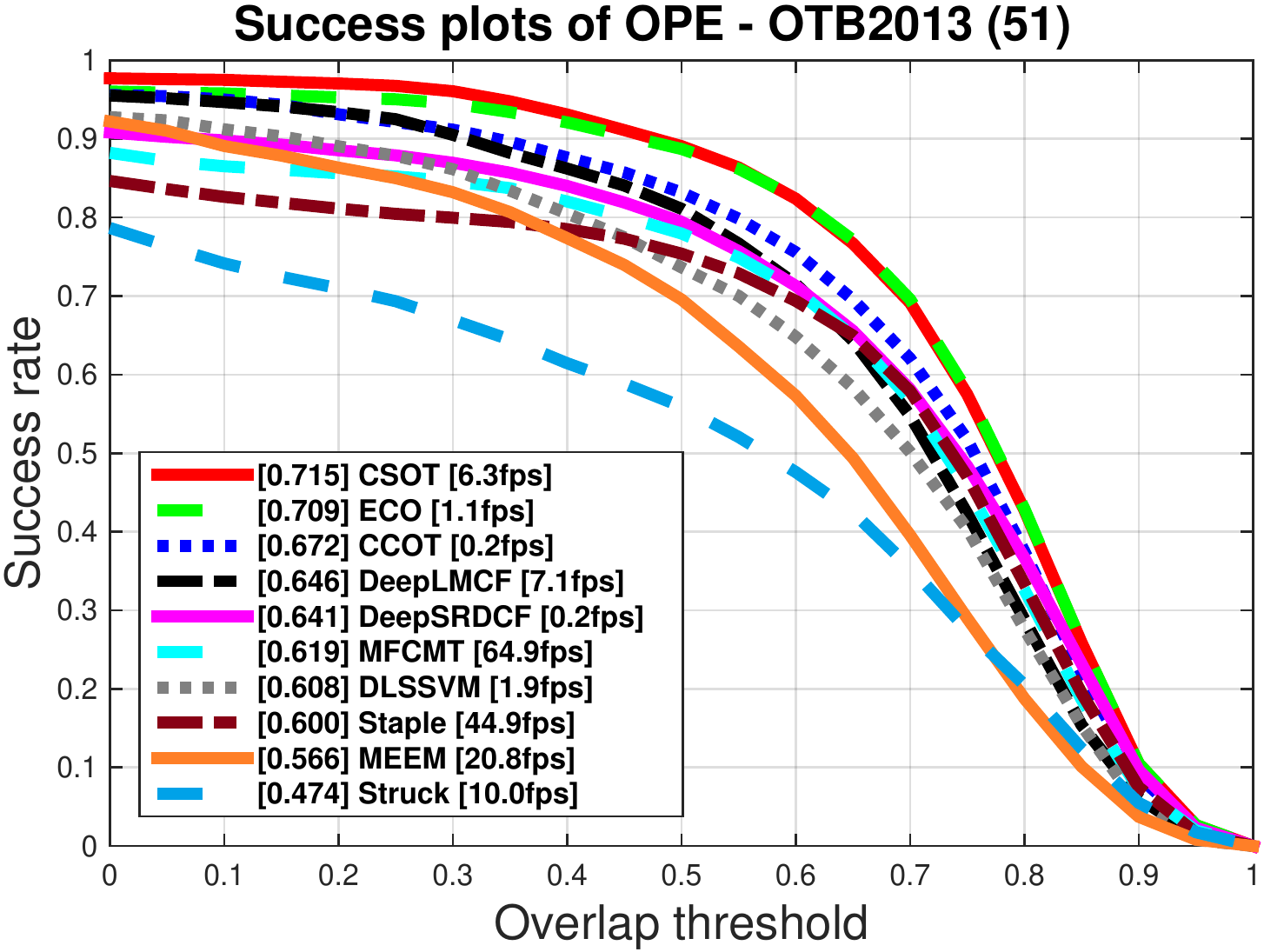}}\end{minipage}
\hfill\begin{minipage}{0.32\linewidth}\centerline{\includegraphics[width=\textwidth]{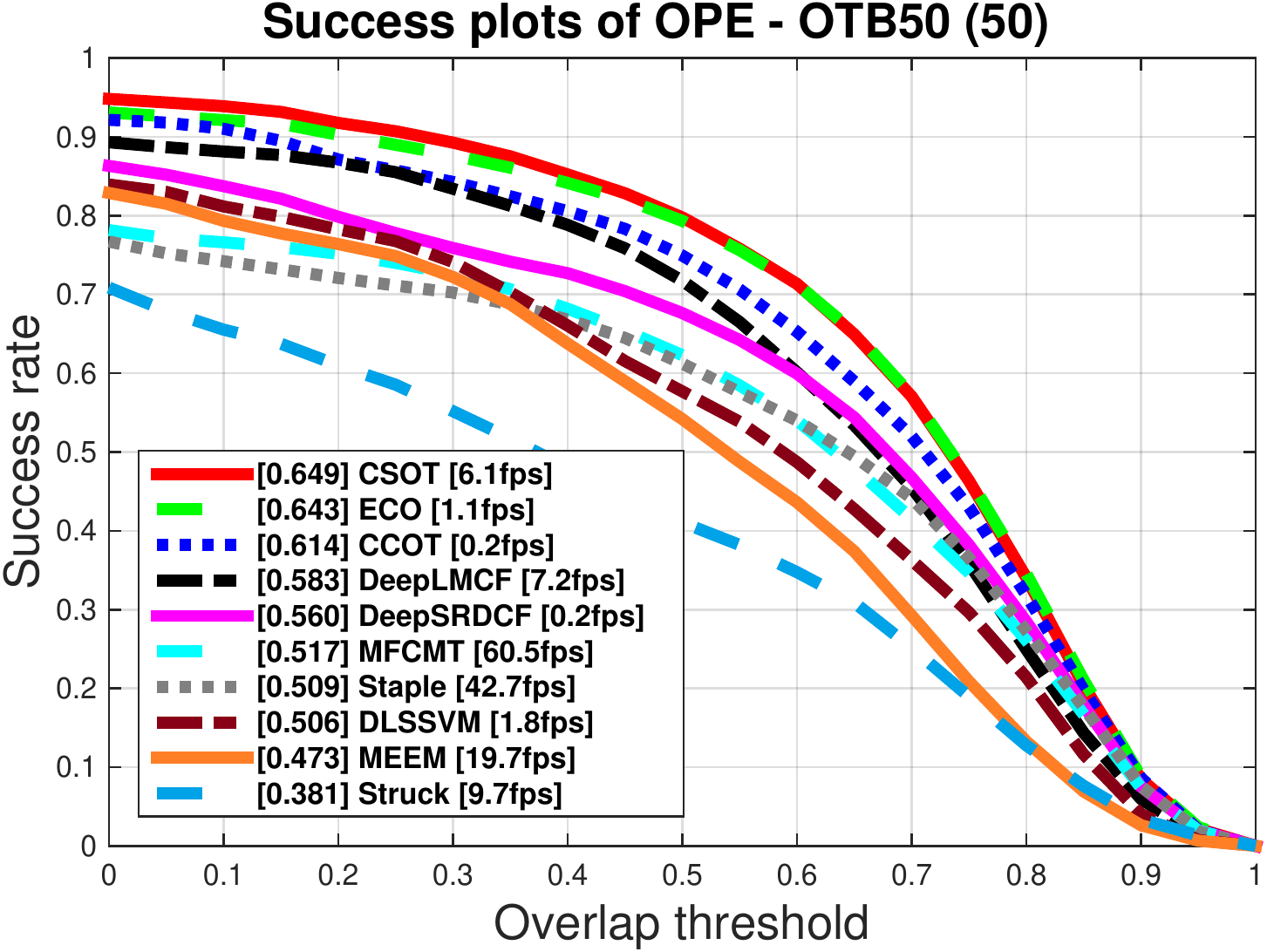}}\end{minipage}
\vspace{1em}
\vfill\begin{minipage}{0.32\linewidth}\centerline{\includegraphics[width=\textwidth]{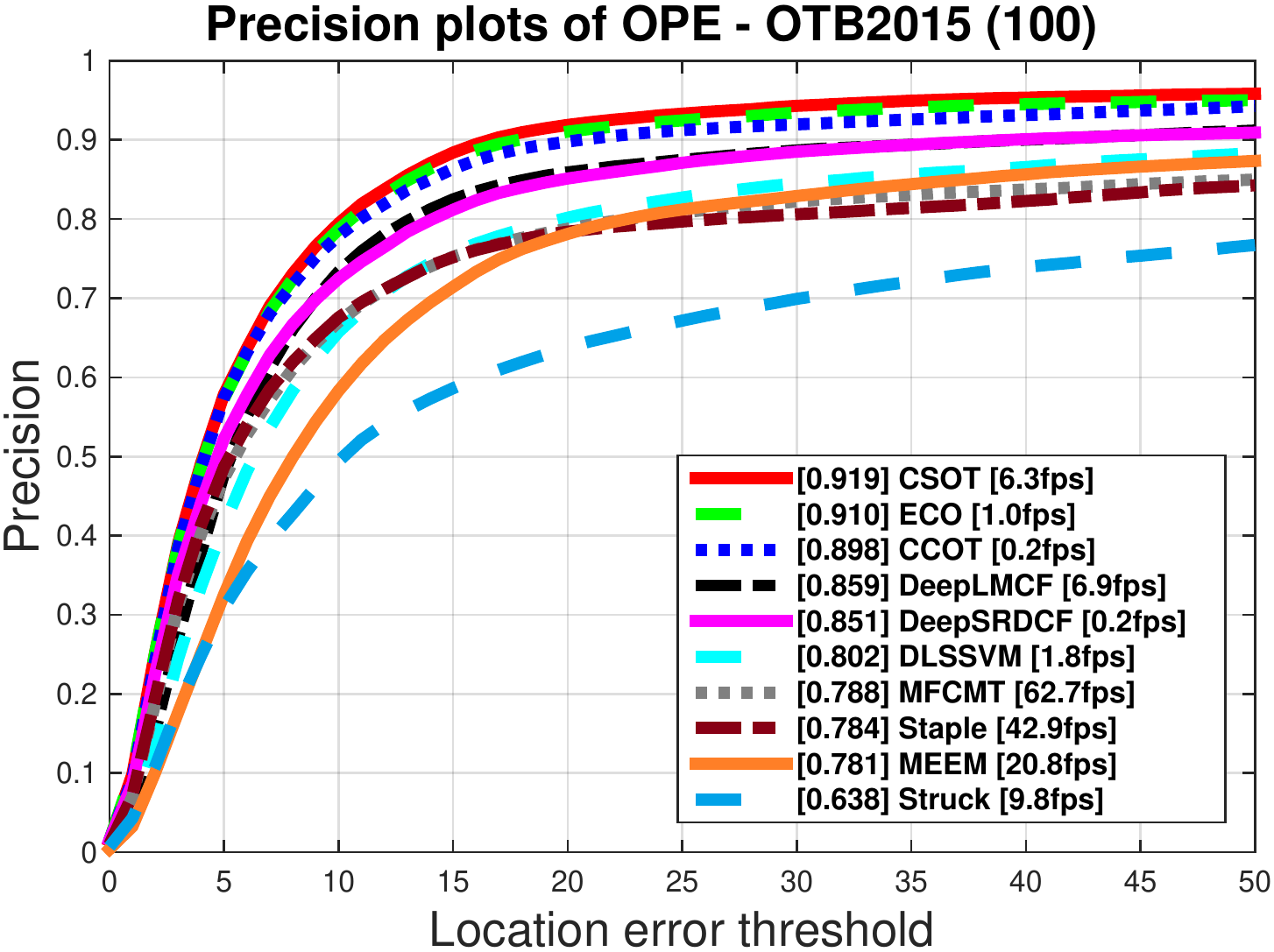}}\end{minipage}
\hfill\begin{minipage}{0.32\linewidth}\centerline{\includegraphics[width=\textwidth]{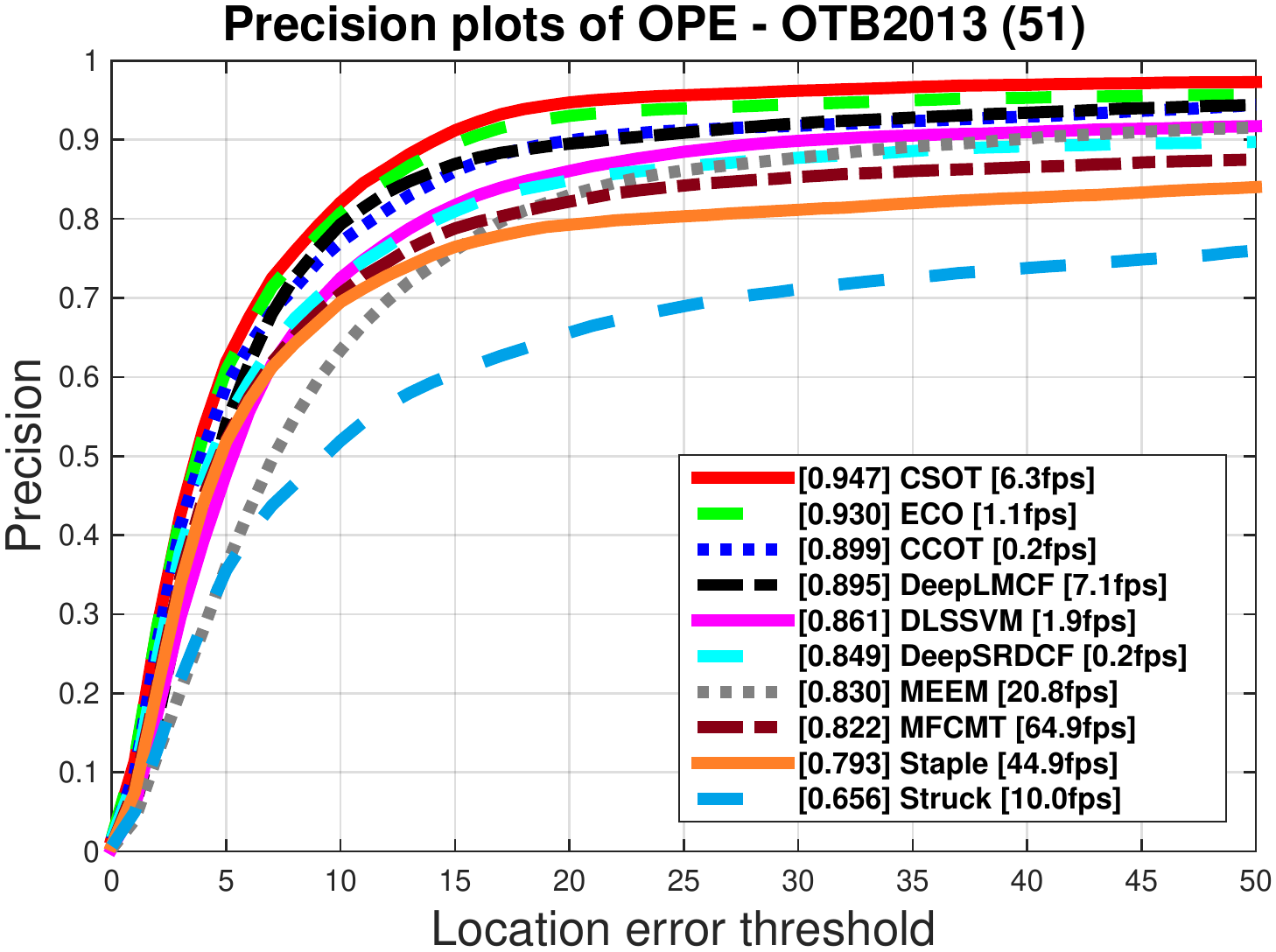}}\end{minipage}
\hfill\begin{minipage}{0.32\linewidth}\centerline{\includegraphics[width=\textwidth]{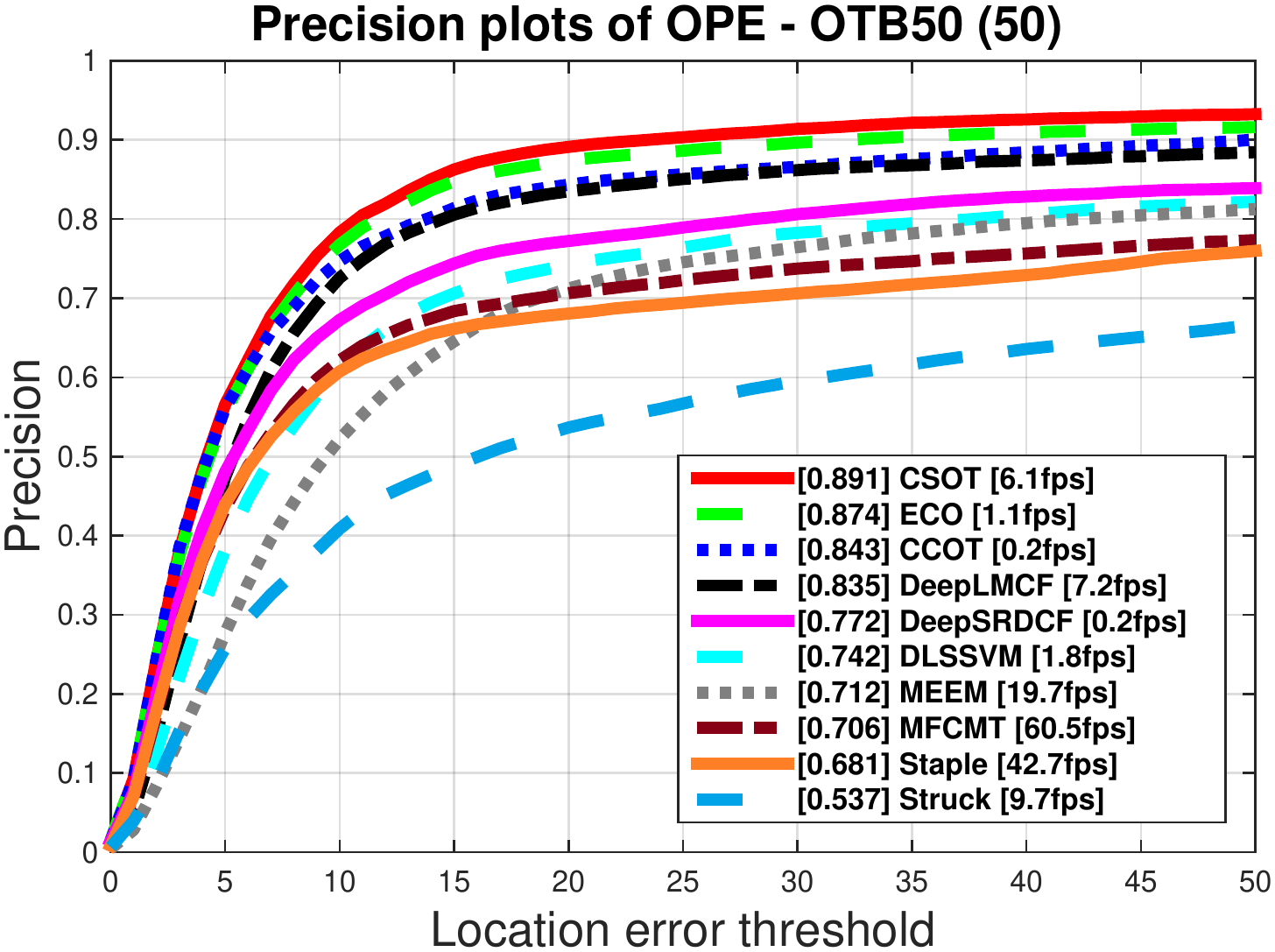}}\end{minipage}
\vspace{-1em}
\caption{Success plots~(top row) and precision plots~(bottom row) of CSOT and nine state-of-the-art trackers over three OTB benchmarks. The first value in the legend indicates the AUC or DP score for each tracker. We also show the speed of the trackers in mean FPS as the last number in the legend. Best viewed in color.}
\label{fig:4}
\end{figure}

\indent We compare CSOT with nine state-of-the-art trackers: ECO~\cite{eco}, C-COT~\cite{ccot}, DeepLMCF~\cite{lmcf}, DeepSRDCF~\cite{deepsrdcf}, DLSSVM~\cite{dlssvm}, MEEM~\cite{meem}, Staple~\cite{staple}, MFCMT~\cite{mfcmt} and Struck~\cite{struck} on OTB50/2013/2015 benchmarks. Among them, DeepLMCF and DeepSRDCF use deep appearance features as a substitute to conventional handcrafted features employed by LMCF and SRDCF. Struck, DLSSVM, DeepLMCF and MEEM are all SOSVM-based methods. ECO and C-COT are the most popular approaches based on correlation filters and continuous convolution operations. It is worth to mention that MEEM, Staple and MFCMT are three different ensemble-based trackers. The success plots and precision plots of all participating trackers are shown in Fig.~\ref{fig:4}.

\indent The top row of Fig.~\ref{fig:4} presents success plots of all participating trackers. Struck which is the best tracker in the original OTB benchmark~\cite{otb2013} obtains a mean AUC score of 43.9\%. MEEM provides a mean AUC score of 52.3\%. DLSSVM further improves the mean AUC score to 55.8\%. Ensemble-based trackers which employing conventional handcrafted features, i.e., Staple and MFCMT, achieve mean AUC scores of 56.3\% and 57.2\%, respectively. The deep features-based versions of SRDCF and LMCF, namely DeepSRDCF and DeepLMCF, have gained improvements of 4.0\% and 4.9\% compared to MFCMT, respectively. By employing continuous convolution, C-COT achieves a mean AUC score of 65.2\%. ECO provides the best results with a mean AUC score of 68.1\%. CSOT obtains state-of-the-art results with a mean AUC score of 68.6\% on OTB benchmarks, which is a significant gain of 0.5\% over the mean AUC score of ECO. The precision plots of these ten trackers are shown in the bottom row of Fig.~\ref{fig:4}. Among them, ECO provides the best performance with a mean DP score of 90.4\%. Our CSOT approach significantly outperforms ECO and obtains a state-of-the-art result with an absolute gain of 1.5\% in mean DP score. Moreover, ECO and C-COT can only run at speeds of 1.1 FPS and 0.2 FPS while our CSOT is superior to them in all evaluations. Interestingly, we notice that CSOT-HC which implemented based on conventional handcrafted features obtains a mean AUC score of 61.9\%. It outperforms most existing trackers, even DeepLMCF and DeepSRDCF. All these experimental results illustrate that our proposed approach is effective and achieves appealing performance.
\begin{table}[t]
\centering\scriptsize
\caption{Attribute-based comparison of CSOT with nine state-of-the-art trackers on the OTB2015 benchmark~\cite{otb2015}. The results are presented in terms of \emph{AUC scores (\%)}. Top three results are highlighted in \textcolor[rgb]{1.00,0.00,0.00}{\textbf{red}}, \textcolor[rgb]{0.00,0.00,1.00}{\textbf{blue}} and \textcolor[rgb]{0.00,1.00,0.00}{\textbf{green}}, respectively. Best viewed in color.}
\vspace{-1em}
\begin{tabular}{p{2.6cm}p{1.2cm}<{\centering}p{1.2cm}<{\centering}p{1.2cm}<{\centering}p{1.4cm}<{\centering}p{1.4cm}<{\centering}p{1.2cm}<{\centering}
p{1.2cm}<{\centering}p{1.2cm}<{\centering}}
\toprule
Attribute & CSOT(ours) & ECO~\cite{eco} & CCOT~\cite{ccot} & DeepSRDCF~\cite{deepsrdcf} & DeepLMCF~\cite{lmcf} & DLSSVM~\cite{dlssvm} &  MEEM~\cite{meem} & Struck~\cite{struck} \\ \midrule
Fast motion (39)
& \textcolor[rgb]{0.00,0.00,1.00}{\textbf{67.8}}
& \textcolor[rgb]{1.00,0.00,0.00}{\textbf{68.3}}
& \textcolor[rgb]{0.00,1.00,0.00}{\textbf{67.5}}
& 62.8 & 60.4 & 53.9 & 54.2 & 46.7 \\
Background clutter (31)
& \textcolor[rgb]{0.00,0.00,1.00}{\textbf{69.2}}
& \textcolor[rgb]{1.00,0.00,0.00}{\textbf{70.0}}
& \textcolor[rgb]{0.00,1.00,0.00}{\textbf{64.5}}
& 62.7 & 60.8 & 54.7 & 51.9 & 43.3 \\
Motion blur (29)
& \textcolor[rgb]{0.00,0.00,1.00}{\textbf{70.7}}
& \textcolor[rgb]{1.00,0.00,0.00}{\textbf{70.9}}
& \textcolor[rgb]{0.00,1.00,0.00}{\textbf{70.3}}
& 64.2 & 62.7 & 57.6 & 55.6 & 46.3 \\
Deformation (44)
& \textcolor[rgb]{1.00,0.00,0.00}{\textbf{64.1}}
& \textcolor[rgb]{0.00,0.00,1.00}{\textbf{63.3}}
& \textcolor[rgb]{0.00,1.00,0.00}{\textbf{61.5}}
& 56.6 & 56.2 & 51.3 & 48.9 & 38.3 \\
Illumination variation (38)
& \textcolor[rgb]{1.00,0.00,0.00}{\textbf{72.2}}
& \textcolor[rgb]{0.00,0.00,1.00}{\textbf{71.3}}
& \textcolor[rgb]{0.00,1.00,0.00}{\textbf{67.6}}
& 62.1 & 62.5 & 56.1 & 51.7 & 42.0 \\
In-plane rotation (51)
& \textcolor[rgb]{1.00,0.00,0.00}{\textbf{66.2}}
& \textcolor[rgb]{0.00,0.00,1.00}{\textbf{65.5}}
& \textcolor[rgb]{0.00,1.00,0.00}{\textbf{62.3}}
& 58.9 & 59.8 & 55.6 & 42.9 & 45.1 \\
Low resolution (9)
& \textcolor[rgb]{0.00,0.00,1.00}{\textbf{61.3}}
& \textcolor[rgb]{0.00,1.00,0.00}{\textbf{59.1}}
& \textcolor[rgb]{1.00,0.00,0.00}{\textbf{62.5}}
& 56.1 & 56.8 & 43.6 & 38.2 & 31.3 \\
Occlusion (49)
& \textcolor[rgb]{0.00,1.00,0.00}{\textbf{62.7}}
& \textcolor[rgb]{1.00,0.00,0.00}{\textbf{68.0}}
& \textcolor[rgb]{0.00,0.00,1.00}{\textbf{67.0}}
& 60.1 & 58.9 & 53.3 & 50.4 & 39.1 \\
Out-of-plane rotation (63)
& \textcolor[rgb]{0.00,1.00,0.00}{\textbf{63.6}}
& \textcolor[rgb]{1.00,0.00,0.00}{\textbf{67.3}}
& \textcolor[rgb]{0.00,0.00,1.00}{\textbf{64.9}}
& 60.7 & 60.5 & 54.7 & 52.5 & 42.4 \\
Out of view (14)
& \textcolor[rgb]{0.00,0.00,1.00}{\textbf{65.8}}
& \textcolor[rgb]{1.00,0.00,0.00}{\textbf{66.0}}
& \textcolor[rgb]{0.00,1.00,0.00}{\textbf{64.5}}
& 55.3 & 60.8 & 46.7 & 48.8 & 37.4 \\
Scale variation (64)
& \textcolor[rgb]{1.00,0.00,0.00}{\textbf{67.1}}
& \textcolor[rgb]{0.00,0.00,1.00}{\textbf{66.6}}
& \textcolor[rgb]{0.00,1.00,0.00}{\textbf{65.1}}
& 60.5 & 58.8 & 48.8 & 47.0 & 40.2 \\\bottomrule
\end{tabular}
\label{table:1}
\end{table}

\indent For more detailed comparisons, we perform attribute-based analysis of CSOT and nine state-of-the-art trackers on the OTB2015 benchmark. Video sequences contained in OTB benchmarks are annotated with 11 different attributes that represent a variety of challenging factors including fast motion, background clutter, motion blur, deformation, illumination variation, in-plane rotation, low resolution, occlusion, out-of-plane rotation, out of view and scale variation. The results of AUC scores are summarized in Table~\ref{table:1}. It is clear that with the exception of occlusions and out-of-plane rotations, CSOT obtains top two results compared to other trackers on nine out of eleven attributes. Moreover, CSOT achieves significant improvements in four scenarios compared to the best existing method and provides the best performance: deformations~(0.8\%), illumination variations~(0.9\%), in-plane rotations~(0.7\%) and scale variations~(0.5\%). These experimental evaluations illustrate that the combination of multiple deep features and the ensemble post-processor have discriminative capabilities superior to other approaches.
\subsection{Experiments on VOT}
The VOT challenge is the largest annual competition in the field of visual tracking. We use the VOT2017 benchmark~\cite{vot2017} which contains 60 video sequences to evaluate our proposed CSOT and eight state-of-the-are trackers, including C-COT~\cite{ccot}, CFCF~\cite{cfcf}, CFWCR~\cite{cfwcr}, CSRDCF~\cite{csrdcf}, ECO~\cite{eco}, LSART~\cite{lsart}, MCCT~\cite{vot2017} and SiamDCF~\cite{vot2017}. For fair comparisons, we use the original results provided by the VOT challenge committee~\footnote{\url{http://www.votchallenge.net/vot2017/}}.
\begin{table}[tb]
\small
\centering
\caption{Comparison of state-of-the-art trackers on the VOT2017 benchmark~\cite{vot2017}. Performance results are presented in terms of \emph{EAO}, \emph{Accuracy} and \emph{Robustness}. Top three results of each metric are highlighted in \textcolor[rgb]{1.00,0.00,0.00}{\textbf{red}}, \textcolor[rgb]{0.00,0.00,1.00}{\textbf{blue}} and \textcolor[rgb]{0.00,1.00,0.00}{\textbf{green}} respectively.}
\begin{tabular}{p{2.5cm}p{2cm}<{\centering}p{2cm}<{\centering}p{2cm}<{\centering}p{2cm}<{\centering}}
\toprule
Tracker                & EAO & Accuracy & Robustness \\ \midrule
C-COT~\cite{ccot}      & 0.267 & 0.493 & 1.315 \\
CFCF~\cite{cfcf}       & 0.286 & \textcolor[rgb]{0.00,0.00,1.00}{\textbf{0.509}} & 1.169 \\
CFWCR~\cite{cfwcr}     & \textcolor[rgb]{0.00,0.00,1.00}{\textbf{0.303}} & 0.484 & 1.210 \\
CSOT (ours)             & \textcolor[rgb]{0.00,1.00,0.00}{\textbf{0.298}} & 0.498 & \textcolor[rgb]{0.00,0.00,1.00}{\textbf{1.089}} \\
CSRDCF~\cite{csrdcf}   & 0.256 & 0.488 & 1.309 \\
ECO~\cite{eco}         & 0.281 & 0.483 & \textcolor[rgb]{0.00,1.00,0.00}{\textbf{1.117}} \\
LSART~\cite{lsart}     & \textcolor[rgb]{1.00,0.00,0.00}{\textbf{0.323}} & 0.493 & \textcolor[rgb]{1.00,0.00,0.00}{\textbf{0.943}} \\
MCCT~\cite{vot2017}    & 0.270 & \textcolor[rgb]{1.00,0.00,0.00}{\textbf{0.525}} & 1.126 \\
SiamDCF~\cite{vot2017} & 0.250 & \textcolor[rgb]{0.00,1.00,0.00}{\textbf{0.500}} & 1.866 \\ \bottomrule
\end{tabular}
\label{table:2}
\end{table}
\begin{figure}[htb]
\captionsetup{belowskip=0em}
\centering
\begin{minipage}{0.49\linewidth}\centerline{\includegraphics[width=\textwidth]{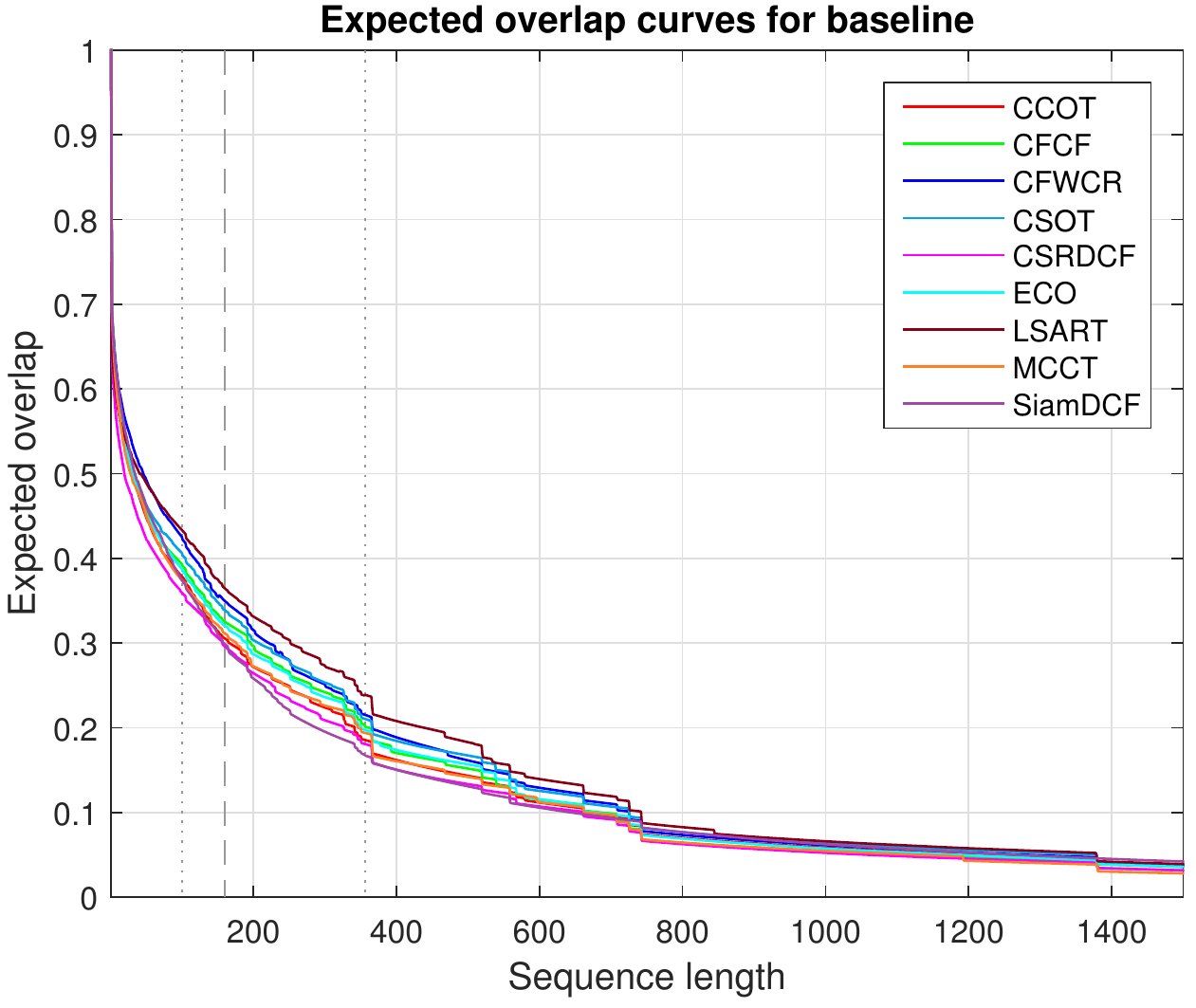}}\end{minipage}
\hfill\begin{minipage}{0.49\linewidth}\centerline{\includegraphics[width=\textwidth]{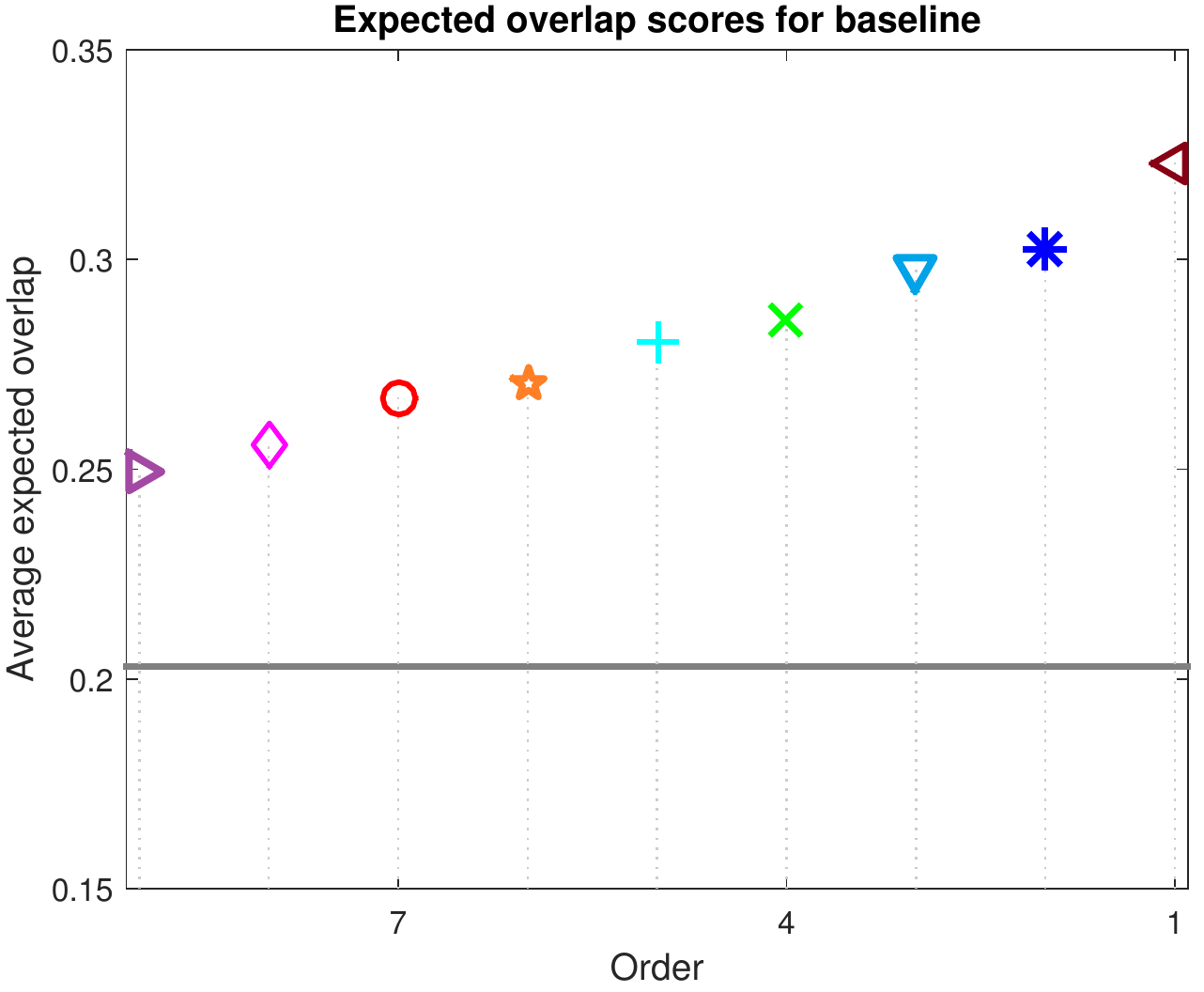}}\end{minipage}
\vspace{1em}
\vfill\begin{minipage}[b]{0.8\linewidth}\centerline{\includegraphics[width=\textwidth]{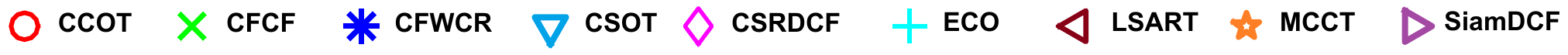}}\medskip\end{minipage}
\vspace{-1em}
\caption{EAO curve~(left)~and EAO graph~(right)~of all participating trackers for experiment baseline on the VOT2017 benchmark~\cite{vot2017}. In the EAO graph, the better performance a tracker achieves, the closer to the right of the graph. The horizontal dashed line denotes the state-of-the-art bound in VOT2017 challenge. Best viewed in color.}
\label{fig:5}
\end{figure}

\indent To evaluate overall performance of all participating trackers, we adopt the EAO curve and EAO graph to visualize tracking performance as shown in Fig.~\ref{fig:5}. The EAO curve demonstrates the average of per-frame overlaps, including the zero overlaps after failures, on different lengths of long sequences. The EAO graph illustrates the EAO score of each tracker, which is computed as the average of the EAO curve values over an interval $[108,\,371]$~\cite{vot2015} of typical short-term sequence lengths. In the Expected Average Overlap graph, the gray horizontal line indicates the state-of-the-art bound suggested by the VOT2017 challenge committee~\cite{vot2017}.

\indent Table~\ref{table:2} shows comparison results on the VOT2017 benchmark. Among compared trackers, LSART achieves the most favorable results in terms of EAO and robustness, but its accuracy is unsatisfactory. MCCT achieves the best accuracy of 52.5\%. It is clear that CSOT is ranked as the third-best with an EAO score of 29.8\%. Compared with ECO and C-COT, CSOT achieves absolute gains of 1.7\% and 2.9\%, respectively. Moreover, CSOT obtains the second-best robustness of 1.089 and the fourth-best accuracy of 49.8\%.
\begin{figure}[tb]
\captionsetup{belowskip=0em}
\centering
\begin{minipage}{0.32\linewidth}\centerline{\includegraphics[width=\textwidth]{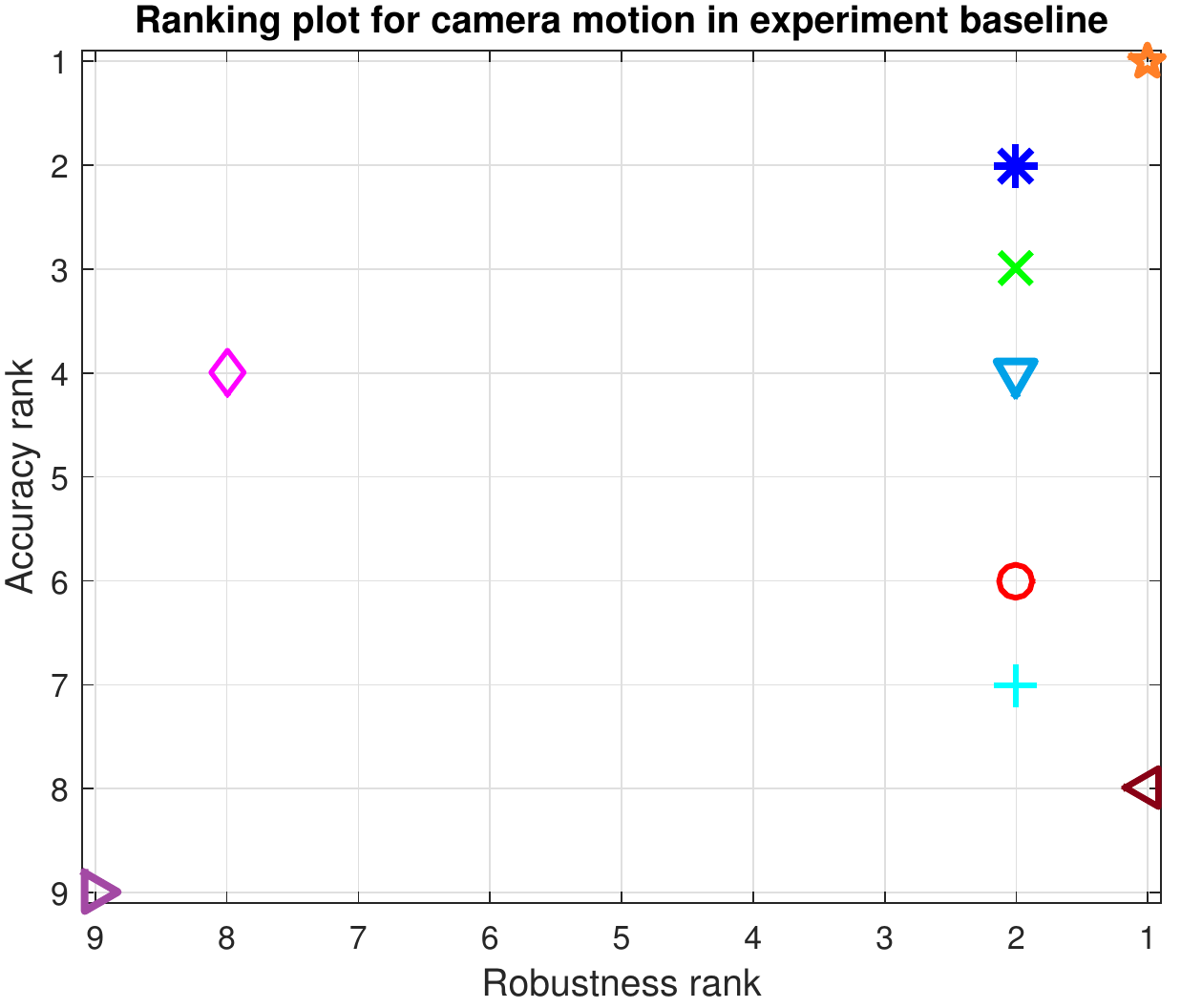}}\end{minipage}
\hfill\begin{minipage}{0.32\linewidth}\centerline{\includegraphics[width=\textwidth]{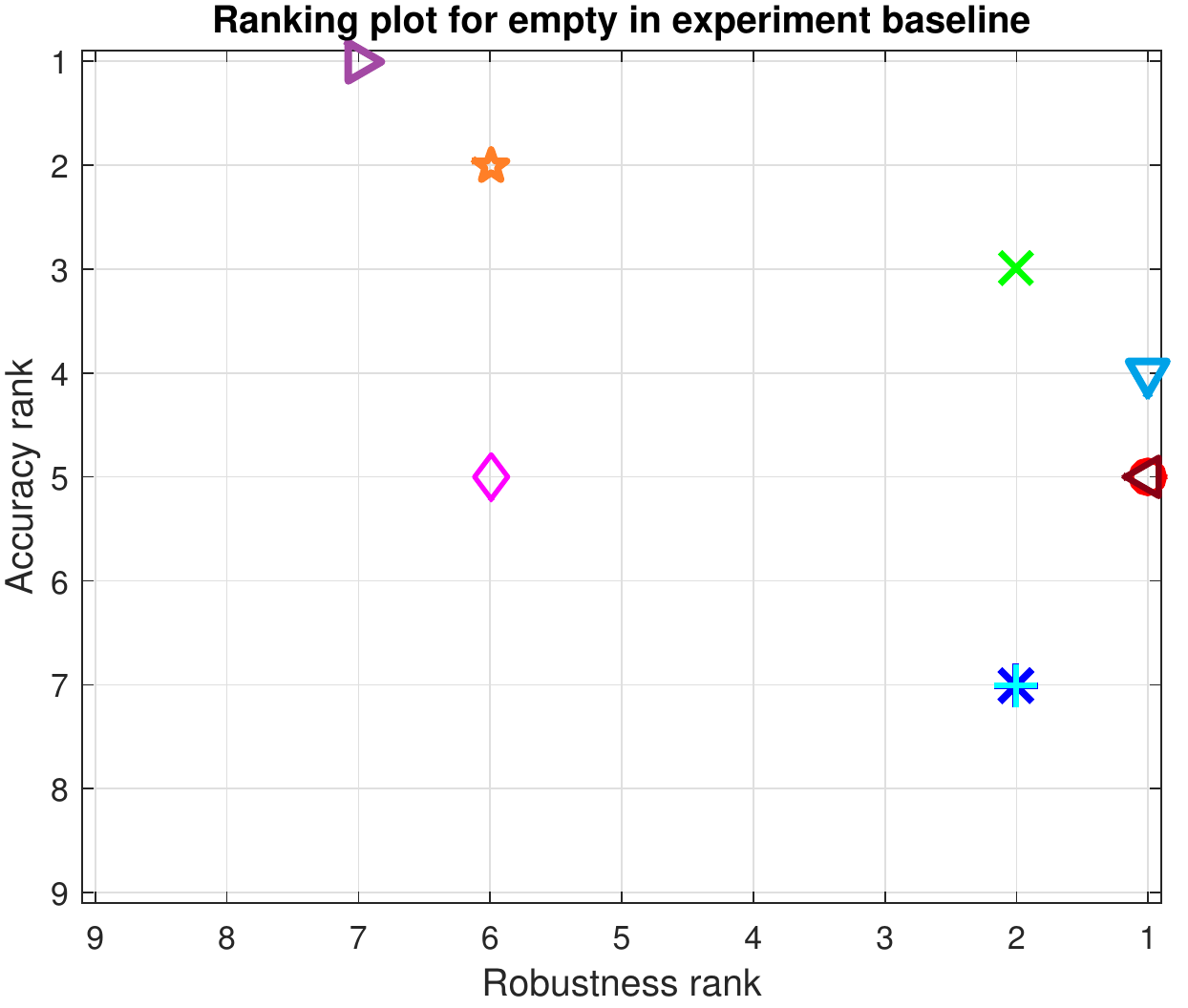}}\end{minipage}
\hfill\begin{minipage}{0.32\linewidth}\centerline{\includegraphics[width=\textwidth]{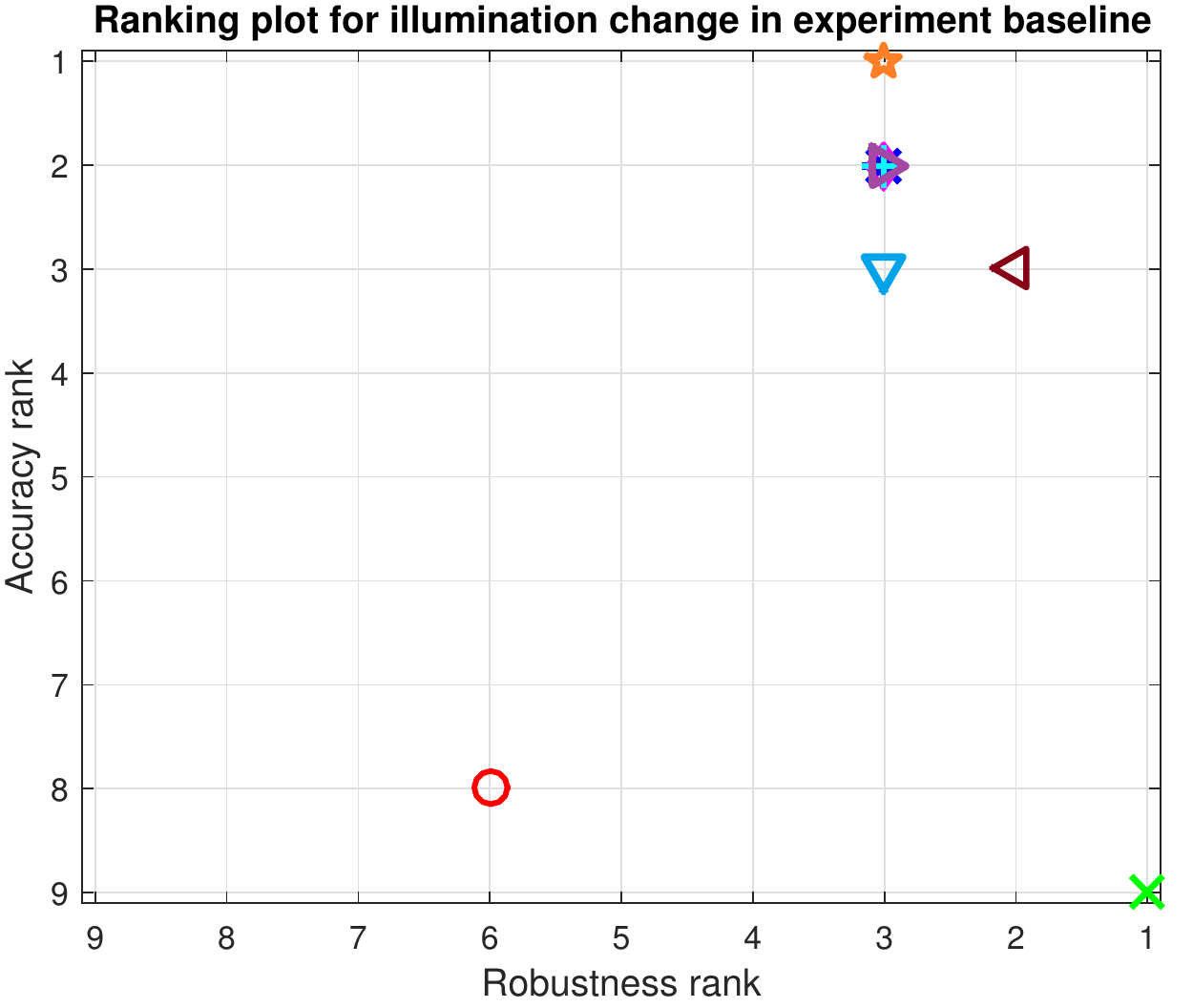}}\end{minipage}
\vspace{1em}
\vfill\begin{minipage}{0.32\linewidth}\centerline{\includegraphics[width=\textwidth]{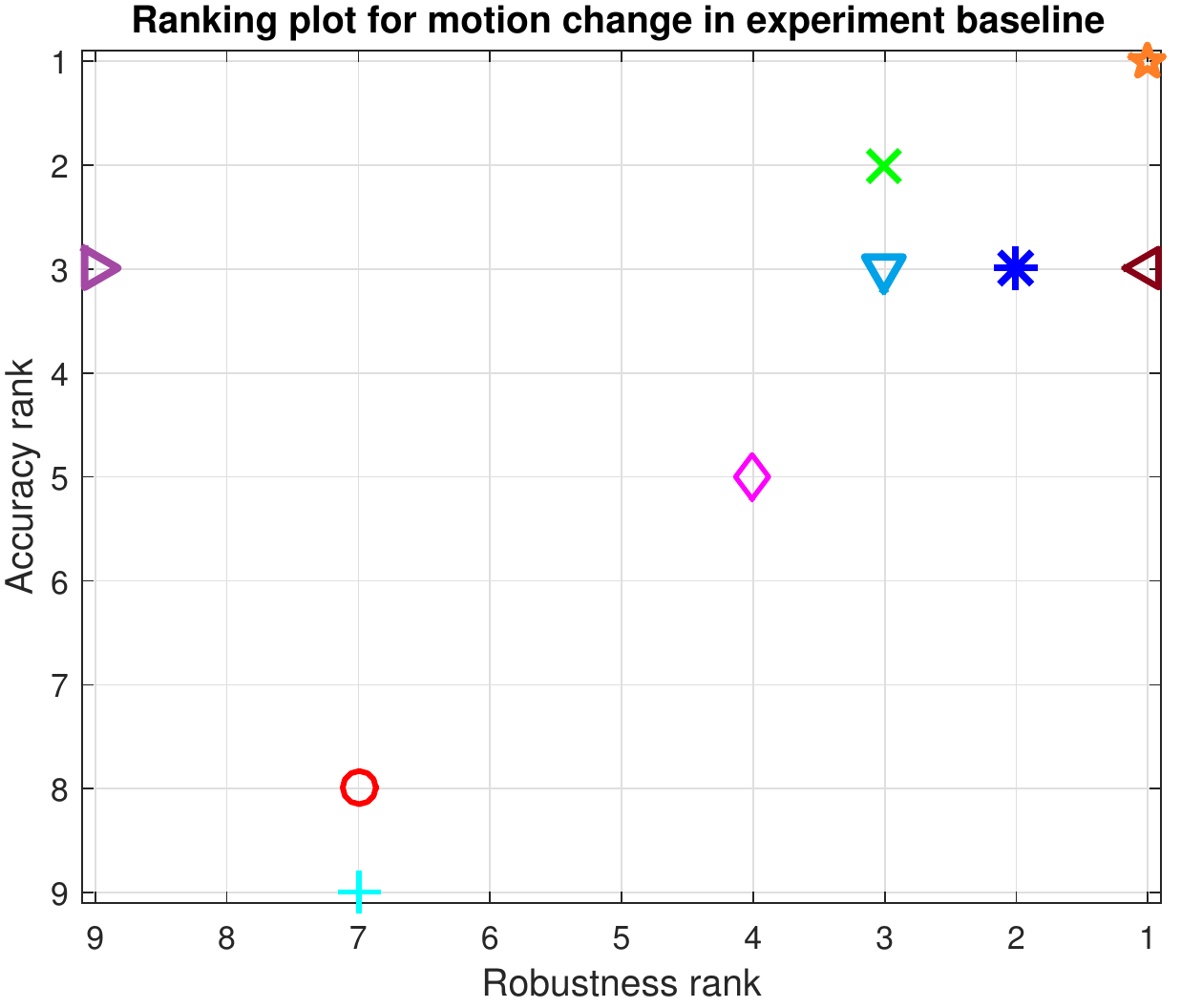}}\end{minipage}
\hfill\begin{minipage}{0.32\linewidth}\centerline{\includegraphics[width=\textwidth]{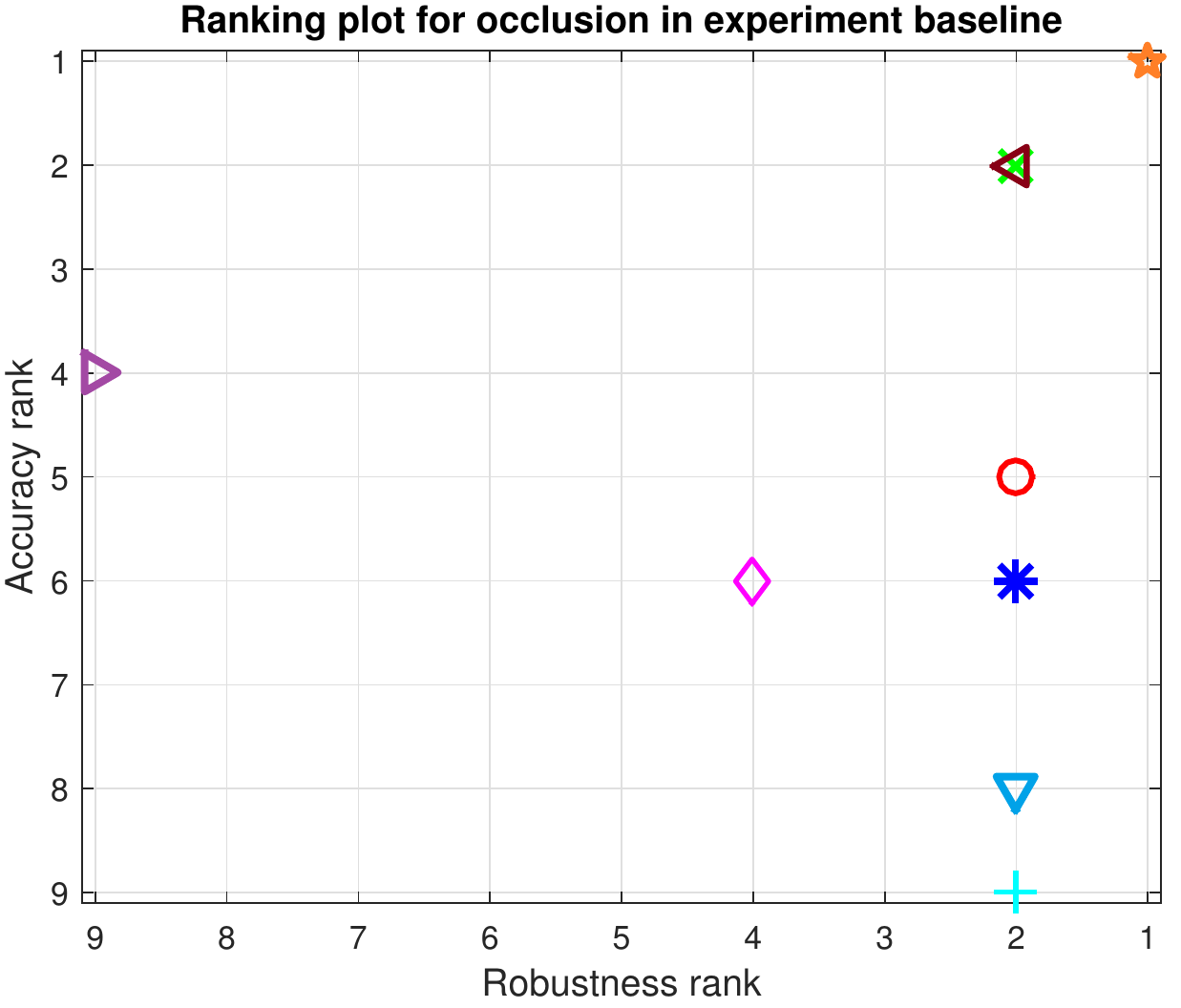}}\end{minipage}
\hfill\begin{minipage}{0.32\linewidth}\centerline{\includegraphics[width=\textwidth]{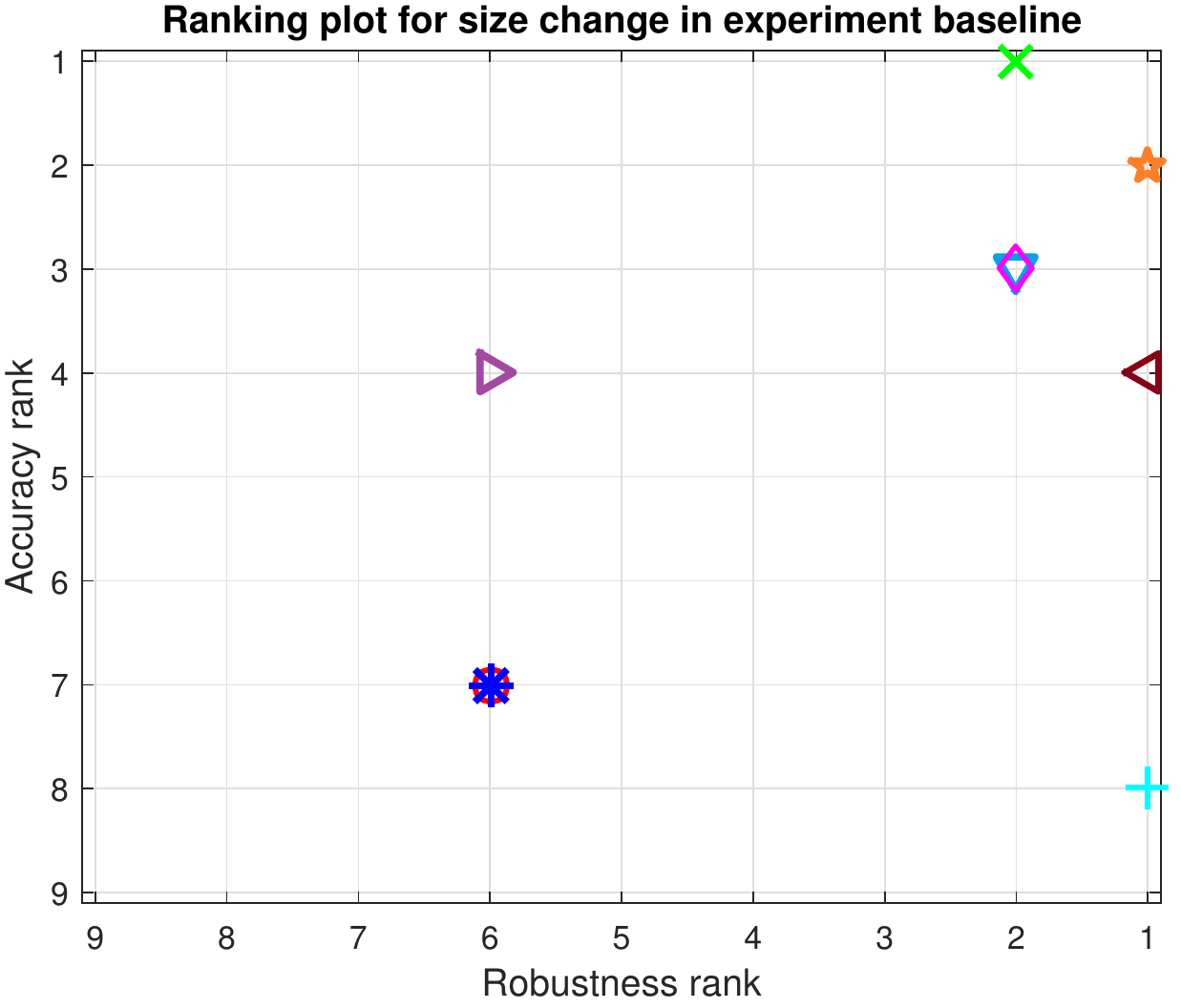}}\end{minipage}
\vspace{1em}
\vfill\begin{minipage}[b]{0.8\linewidth}\centerline{\includegraphics[width=\textwidth]{votlegend.pdf}}\medskip\end{minipage}
\vspace{-1em}
\caption{Ranking plots of the compared trackers in experiment baseline for each attribute (i.e. \emph{camera motion}, \emph{empty}, \emph{illumination change}, \emph{motion change}, \emph{occlusion} and \emph{size change}) on the VOT2017 benchmark~\cite{vot2017}. Here, \emph{empty} denotes frames without labeled attribute. The better performance a tracker achieves, the closer to the top-right corner. Best viewed in color.}
\label{fig:6}
\end{figure}

\indent In the VOT2017 benchmark, all sequences are labeled with five different attributes: camera motion, illumination change, motion change, occlusion and size change. For a more extensive comparison, we further compare each attribute of all participating trackers on the corresponding sub-dataset of VOT2017 benchmark, results are shown in Fig.~\ref{fig:6}. It can be seen that CSOT obtains satisfactory results on all attributes except occlusion. Especially for the empty attribute, our method achieves the best robustness. With regard to camera motion, occlusion and size change challenges, our approach gets the second-best robustness. CSOT also provides the third-best robustness and accuracy with regard to motion change and illumination change.
\begin{figure}[t]
\captionsetup{belowskip=0em}
\centering
\begin{minipage}{0.24\linewidth}\centerline{\includegraphics[width=\textwidth]{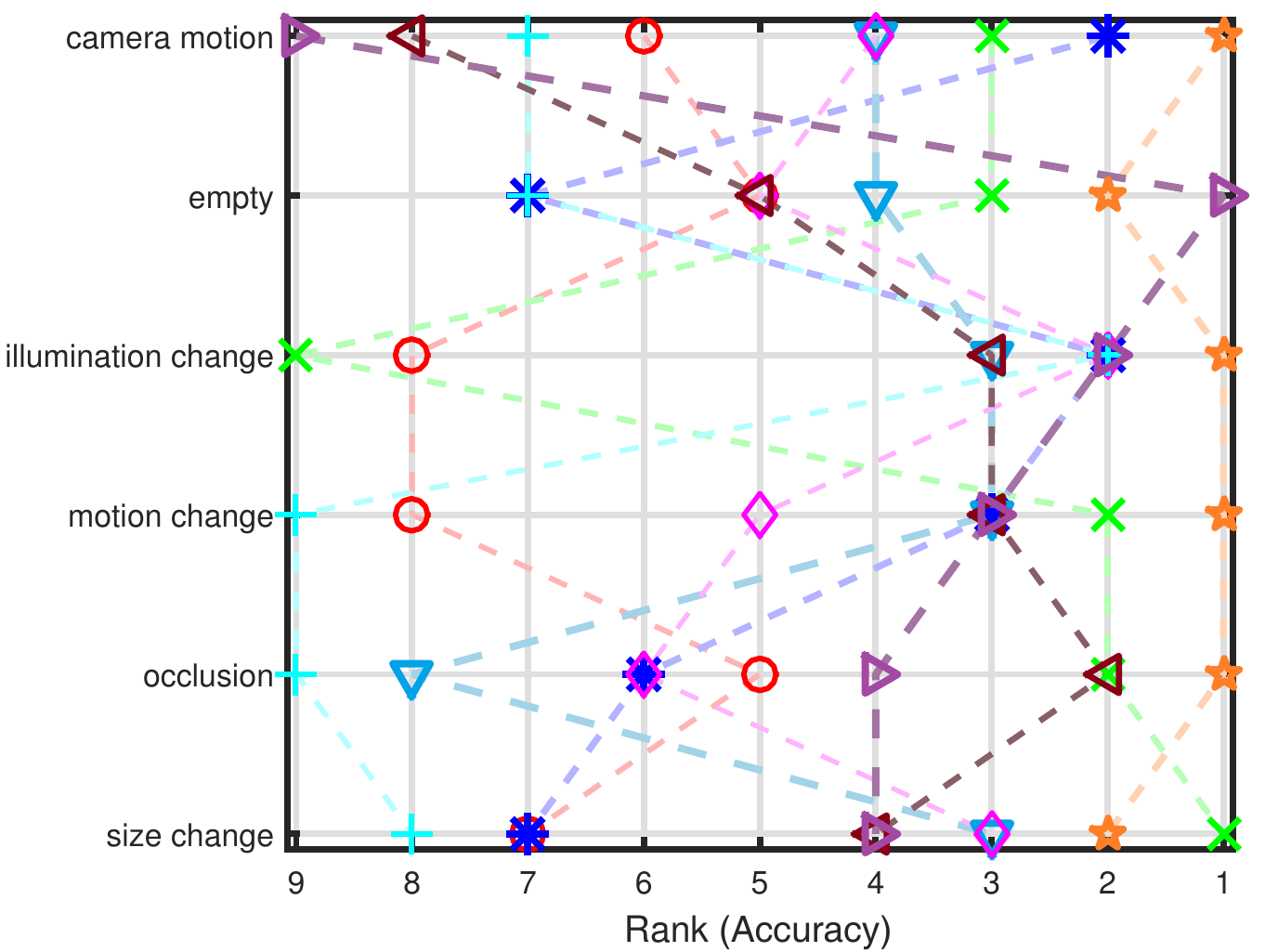}}\end{minipage}
\hfill\begin{minipage}{0.24\linewidth}\centerline{\includegraphics[width=\textwidth]{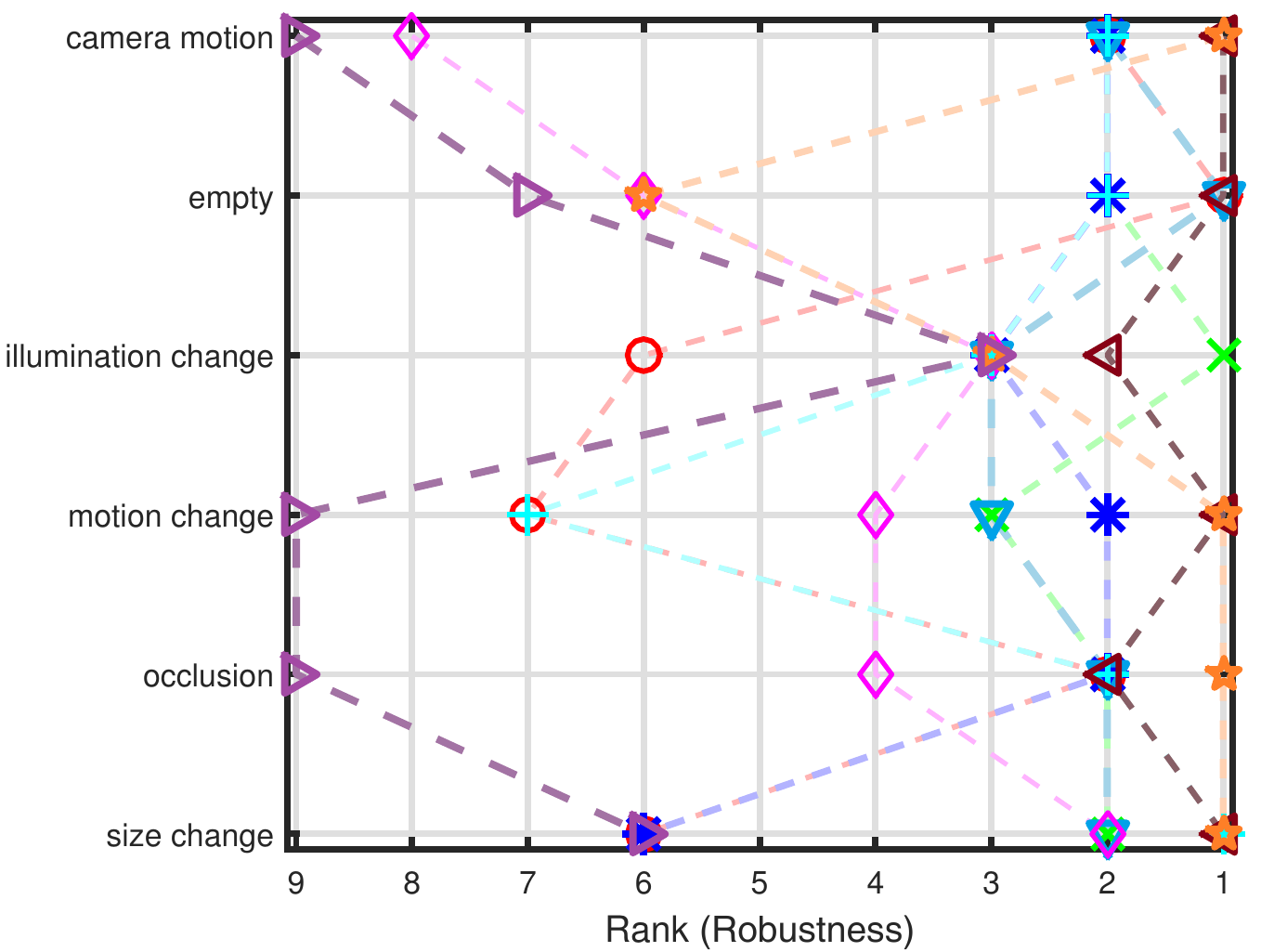}}\end{minipage}
\hfill\begin{minipage}{0.24\linewidth}\centerline{\includegraphics[width=\textwidth]{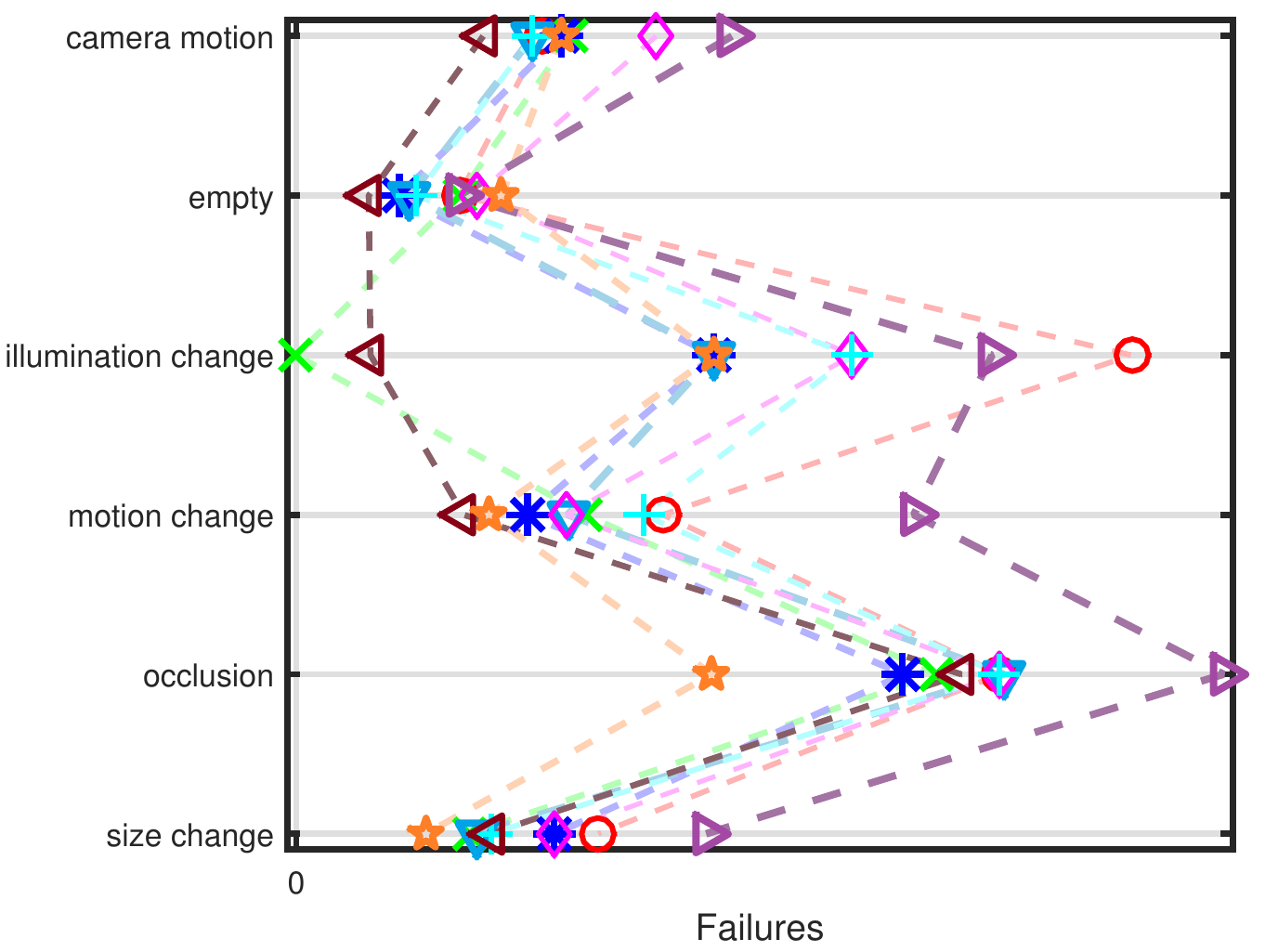}}\end{minipage}
\hfill\begin{minipage}{0.24\linewidth}\centerline{\includegraphics[width=\textwidth]{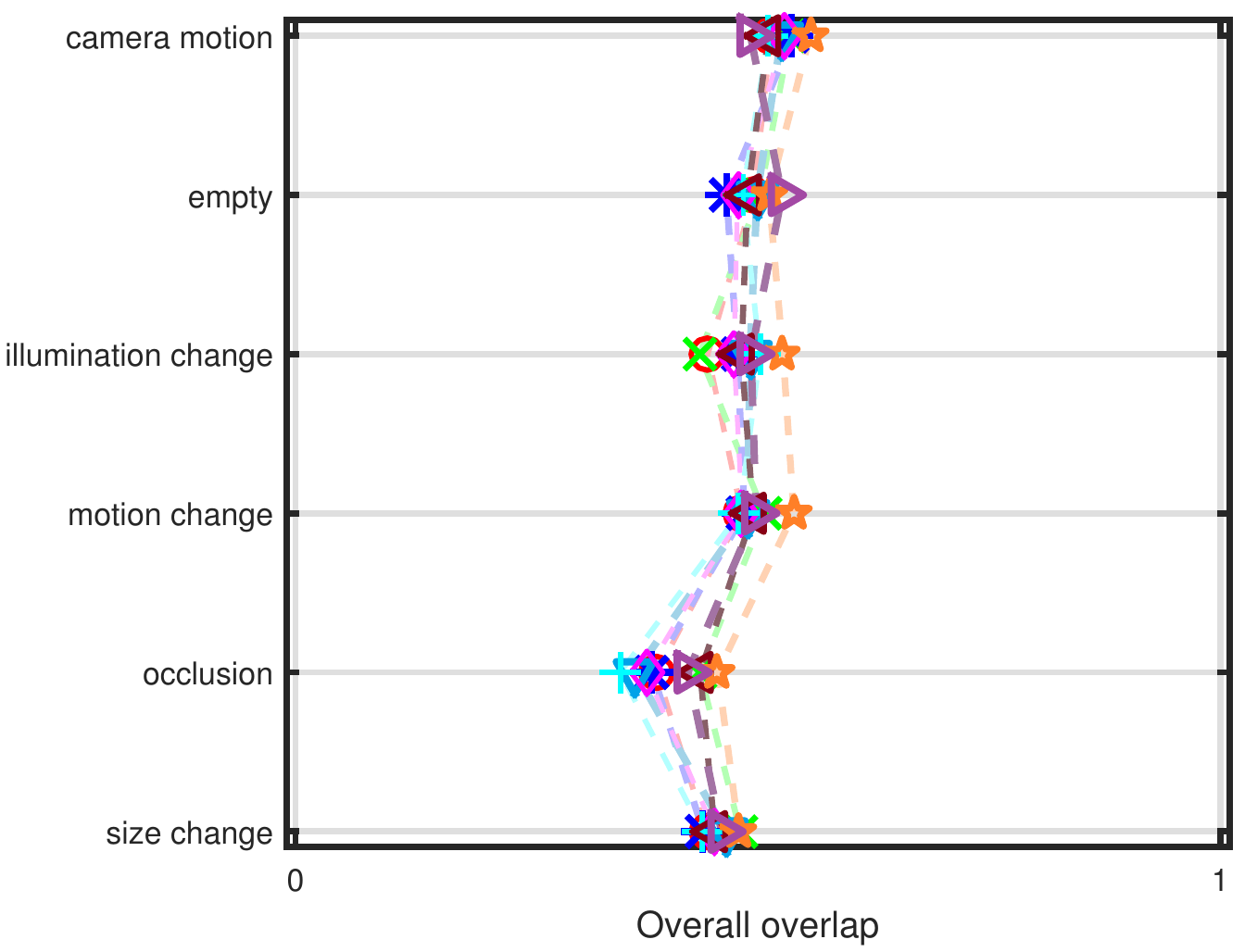}}\end{minipage}
\vspace{1em}
\vfill\begin{minipage}[b]{0.8\linewidth}\centerline{\includegraphics[width=\textwidth]{votlegend.pdf}}\medskip\end{minipage}
\vspace{-1em}
\caption{Ranking of all participating trackers in experiment baseline for each attribute (i.e. \emph{camera motion}, \emph{empty}, \emph{illumination change}, \emph{motion change}, \emph{occlusion} and \emph{size change}) on the VOT2017 benchmark~\cite{vot2017}. Here, \emph{empty} denotes frames without the labeled attribute. The better performance a tracker achieves, the closer to the right of the plot. Best viewed in color.}
\label{fig:7}
\end{figure}

\indent Fig.~\ref{fig:7} individually visualizes the ranking of all participating trackers in terms of accuracy, robustness, failures and overall overlap for each attribute. Our approach achieves promising results on all five attributes. Compared with ECO and C-COT, all metrics are significantly improved. According to Fig.~\ref{fig:7}, it is shown that CSOT is effective and robust to deal with various challenging attributes.
\section{Conclusions} \label{sec:con}

In this paper, we have proposed a novel circular and structural operator tracker (CSOT) for high performance visual tracking. The superior computational efficiency of DCF complemented with the powerful discriminative capability of SOSVM make it possible to employ higher-dimensional deep features and denser circular samples. To obtain the primal confidence score maps, we utilize circular and structural operators to circular correlate deep complementary features with structural correlation filters. The resulting tracker greatly benefits from the heterogeneity of multiple deep features. To improve tracking performance, we propose an ensemble post-processor based on relative entropy, which fuses primal confidence score maps to get an optimal confidence score map. Besides, we utilize an online collaborative optimization strategy to efficiently update circular and structural operators by training structural correlation filters. Experimental evaluations clearly demonstrate that our CSOT outperforms far above most state-of-the-art trackers both in terms of accuracy and robustness on OTB benchmarks, and it also achieves impressive results on the VOT2017 benchmark.
\section*{Acknowledgments} \label{sec:ack}

This work was supported in part by the Science and Technology Planning Program of Guangdong Province, China under Grant 2013B090600105 and Grant 2016B090918047.

The authors would like to thank all the anonymous reviewers for their insightful comments and suggestions that have helped to significantly improve the quality of this paper.
%%
%\end{spacing}
\section*{References} \label{sec:refer}
\bibliography{CSOT}
\end{document}